\documentclass[sigconf]{acmart}
\AtBeginDocument{%
  }

\setcopyright{cc}
\setcctype{by}
\copyrightyear{2026}
\acmYear{2026}
\acmDOI{10.1145/3770855.3817539}
\acmConference[KDD '26]{Proceedings of the 32nd ACM SIGKDD Conference
  on Knowledge Discovery and Data Mining V.2}{August 09--13,
  2026}{Jeju Island, Republic of Korea}
\acmBooktitle{Proceedings of the 32nd ACM SIGKDD Conference on Knowledge Discovery and Data Mining V.2 (KDD '26), August 09--13, 2026, Jeju Island, Republic of Korea}
\acmISBN{979-8-4007-2259-2/2026/08}

\usepackage{microtype}
\usepackage{graphicx}
\usepackage{tcolorbox}
\usepackage{booktabs}
\usepackage{multirow}
\usepackage{tabularx}
\usepackage[table]{xcolor}
\usepackage{makecell}
\usepackage{balance}
\newcommand{\XSH}[1]{{\color{black}#1}}

\begin{document}

\title[A Multi-Agent Framework for Benchmarking LLMs in Chinese Psychiatric Consultation and Diagnosis]{LingxiDiagBench: A Multi-Agent Framework for Benchmarking LLMs in Chinese Psychiatric Consultation and Diagnosis}

\author{Shihao Xu}
\affiliation{%
  \institution{Tianqiao and Chrissy Chen Institute}
  \city{Shanghai}
  \country{China}}
\affiliation{%
  \institution{EverMind AI Inc.}
  \city{California}
  \country{USA}}
\email{shihao.xu@shanda.com}

\author{Tiancheng Zhou}
\affiliation{%
  \institution{EverMind AI Inc.}
  \city{California}
  \country{USA}}

\author{Jiatong Ma}
\affiliation{%
  \institution{EverMind AI Inc.}
  \city{California}
  \country{USA}}

\author{Yanli Ding}
\affiliation{%
  \institution{Shanghai Mental Health Center, Shanghai Jiao Tong University School of Medicine}
  \city{Shanghai}
  \country{China}}

\author{Yiming Yan}
\affiliation{%
  \institution{Shanghai Mental Health Center, Shanghai Jiao Tong University School of Medicine}
  \city{Shanghai}
  \country{China}}

\author{Ming Xiao}
\affiliation{%
  \institution{EverMind AI Inc.}
  \city{California}
  \country{USA}}

\author{Guoyi Li}
\affiliation{%
  \institution{EverMind AI Inc.}
  \city{California}
  \country{USA}}

\author{Haiyang Geng}
\affiliation{%
  \institution{Tianqiao and Chrissy Chen Institute}
  \city{Shanghai}
  \country{China}}
\affiliation{%
  \institution{EverMind AI Inc.}
  \city{California}
  \country{USA}}

\author{Yunyun Han}
\affiliation{%
  \institution{Tianqiao and Chrissy Chen Institute}
  \city{Shanghai}
  \country{China}}
\affiliation{%
  \institution{EverMind AI Inc.}
  \city{California}
  \country{USA}}

\author{Jianhua Chen}
\affiliation{%
  \institution{Shanghai Mental Health Center, Shanghai Jiao Tong University School of Medicine}
  \city{Shanghai}
  \country{China}}

\author{Yafeng Deng}
\affiliation{%
  \institution{EverMind AI Inc.}
  \city{California}
  \country{USA}}

\renewcommand{\shortauthors}{Shihao Xu et al.}

\begin{abstract}
Mental disorders are highly prevalent worldwide, but the shortage of psychiatrists and the inherent subjectivity of interview-based diagnosis create substantial barriers to timely and consistent mental-health assessment.
Progress in AI-assisted psychiatric diagnosis is constrained by the absence of benchmarks that simultaneously provide realistic patient simulation, clinician-verified diagnostic labels, and support for dynamic multi-turn consultation.
We present LingxiDiagBench, a large-scale multi-agent benchmark that evaluates LLMs on both static diagnostic inference and dynamic multi-turn psychiatric consultation in Chinese.
At its core is LingxiDiag-16K, a dataset of 16,000 EMR-aligned synthetic consultation dialogues designed to reproduce real clinical demographic and diagnostic distributions across 12 ICD-10 psychiatric categories.
Through extensive experiments across state-of-the-art LLMs, we establish key findings: (1) although LLMs achieve high accuracy on binary depression--anxiety classification (up to 92.3\%), performance deteriorates substantially for depression--anxiety comorbidity recognition (43.0\%) and 12-way differential diagnosis (28.5\%); (2) dynamic consultation often underperforms static evaluation, indicating that ineffective information-gathering strategies significantly impair downstream diagnostic reasoning; (3) consultation quality assessed by LLM-as-a-Judge shows only moderate correlation with diagnostic accuracy, suggesting that well-structured questioning alone does not ensure correct diagnostic decisions.
We release LingxiDiag-16K and the full evaluation framework to support reproducible research at \url{https://github.com/Lingxi-mental-health/LingxiDiagBench}.
\end{abstract}





\maketitle

\begin{figure*}[!htb]
    \centering
    \includegraphics[width=0.9\textwidth]{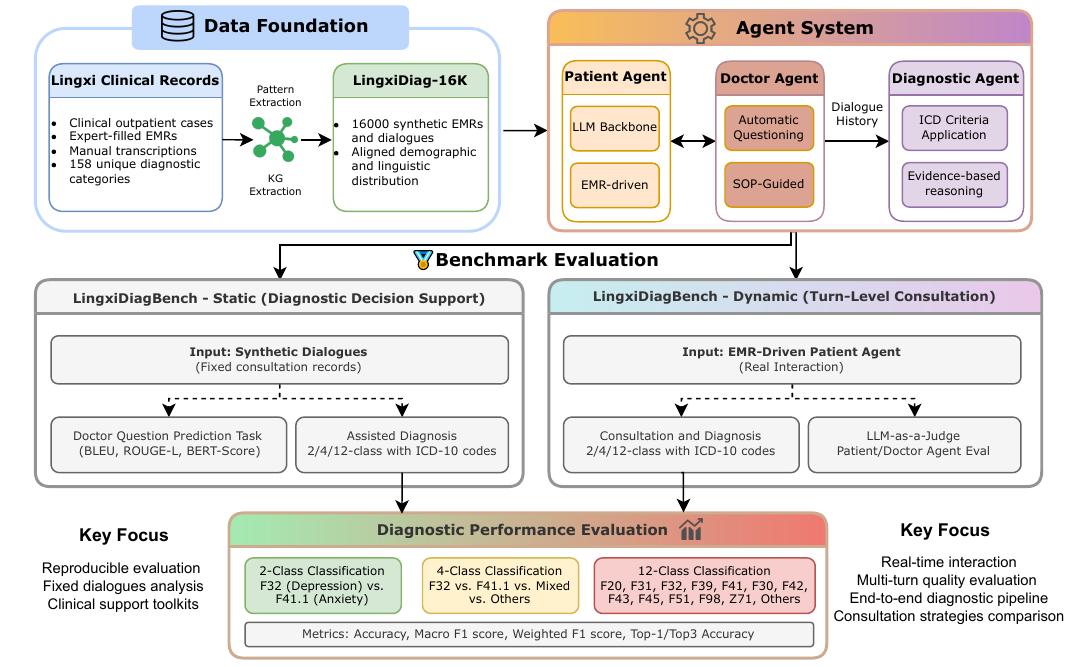}
    \caption{Overview of the LingxiDiagBench framework.}
    \label{fig:framework}
\end{figure*}

\section{Introduction}
\label{sec:intro}

Mental disorders impose a substantial global health burden, affecting roughly one in eight individuals worldwide and placing increasing pressure on mental-health service capacity~\cite{GBD2021}.
The diagnosis of mental disorders relies heavily on clinical interviews, where psychiatrists must synthesize patient-reported symptoms, behavioral observations, and medical history according to standardized criteria~\cite{ICD10Version2016}.
However, the global shortage of mental-health professionals severely restricts timely psychiatric assessment, creating a pressing need for scalable diagnostic-support tools.
Although LLMs have advanced rapidly, current psychiatric-AI benchmarks remain insufficient for evaluating diagnostic capability in realistic clinical scenarios.
Existing benchmarks exhibit three major limitations: (1) most rely on template-based synthetic dialogues that lack realistic conversational variability; (2) they often omit key patient information needed for differential diagnosis and rarely include clinician-verified diagnostic labels; (3) few support dynamic, multi-turn consultation, preventing evaluation of information-gathering strategies.

To address these limitations, we propose LingxiDiagBench, the first large-scale, real-data-driven, multi-disease diagnostic benchmark for psychiatric consultation in Chinese, as shown in Figure~\ref{fig:framework}.
Our main contributions are summarized as follows:
First, we construct LingxiDiag-16K, a dataset of 16,000 synthetic consultation dialogues derived from real electronic medical records and transcripts.
Second, we design three diagnostic tasks with increasing difficulty: binary classification (depression vs. anxiety), four-way classification (including comorbidity of depression and anxiety), and 12-category ICD-10 code multi-label prediction.
Third, we develop an agent-based evaluation framework comprising Patient Agents simulating realistic patient behavior, Doctor Agents implementing diverse consultation strategies, and Diagnosis Agents performing evidence-based diagnosis.
Fourth, we conduct extensive experiments across state-of-the-art LLMs and baseline methods, revealing substantial performance gaps.
Notably, we release the synthetic dataset, consultation framework, and evaluation code. We encourage the community to leverage the framework and benchmark to advance the development of AI-assisted psychiatric diagnosis models, in order to improve access to accurate and timely psychiatric diagnosis in real-world scenarios.

\section{Related Work}
\label{sec:related}

\begin{table*}[t]
\centering
\caption{Comparison of existing benchmarks for mental health and psychiatric AI evaluation. Columns indicate support for real clinical data, synthetic data generation, multi-turn interactive dialogue, agent-based architecture, golden standard criteria, patient agent evaluation, doctor agent evaluation, and diagnostic task.}
\label{tab:benchmark_comparison}
\setlength\tabcolsep{2pt}
\begin{tabular*}{\linewidth}{@{\extracolsep{\fill}}lcccccccc@{}}
\toprule
Benchmark & Clinical Data & Synthetic Data & Interactive & Agent & Golden Standard & Patient Eval & Doctor Eval & Multi-turn Diagnosis  \\
\midrule
PsychiatryBench~\cite{liuPsychiatryBenchComprehensiveLLMEval2024}  &  & $\checkmark$ &  &  & $\checkmark$ &  &  & $\checkmark$ \\
MentalChat16K~\cite{liMentalChat16KBenchmark2025}  &  & $\checkmark$ & $\checkmark$ &  &  &  &  &   \\
Psychosis-Bench~\cite{vargasPsychogenicMachineSimulating2025}  &  & $\checkmark$ & $\checkmark$ &  &  &  &  & \\
MindEval~\cite{ferrazMindEvalMultiTurnMental2025} &   & $\checkmark$ & $\checkmark$ & $\checkmark$ &  & $\checkmark$ & $\checkmark$ & \\
MentraSuite~\cite{xiaoMentraSuitePostTrainingLarge2025b}  & $\checkmark$ & $\checkmark$ &  &  &  &  &  & $\checkmark$  \\
\midrule
\textbf{LingxiDiagBench} & $\checkmark$ & $\checkmark$ & $\checkmark$ & $\checkmark$ & $\checkmark$ & $\checkmark$ & $\checkmark$ & $\checkmark$  \\
\bottomrule
\end{tabular*}
\end{table*}

Recent medical-AI evaluation has undergone a paradigm shift from static knowledge tests toward realistic clinical-dialogue simulation, with growing emphasis on synthetic data construction and end-to-end diagnostic task evaluation~\cite{qiuQuantifyingReasoningAbilities2025,gongMedDialogRubricsComprehensiveBenchmark2026,schmidgallAgentClinicMultimodalAgent2025a}.
Psychiatric AI benchmarks have followed a parallel trajectory; as summarized in Table~\ref{tab:benchmark_comparison}, recent efforts increasingly incorporate data synthesis and diagnostic assessment, yet current benchmarks span diverse but partial evaluation dimensions.
Early efforts such as PsychiatryBench~\cite{liuPsychiatryBenchComprehensiveLLMEval2024} focus on multi-task assessment, including knowledge recall, safety detection, and reasoning through static question-answering, yet they lack interactive consultation capability.
Conversational datasets like MentalChat16K~\cite{liMentalChat16KBenchmark2025} offer dialogue data for mental health assistance but emphasize empathetic response generation rather than diagnostic accuracy.
Safety concerns have also received attention, with Psychosis-Bench~\cite{vargasPsychogenicMachineSimulating2025} specifically testing whether models reinforce delusional content, addressing critical harm-prevention while remaining narrow in diagnostic scope.
Moreover, multi-turn therapeutic dialogue evaluation has been explored in MindEval~\cite{ferrazMindEvalMultiTurnMental2025}, which incorporates both patient and clinician perspectives but prioritizes therapeutic quality over diagnostic precision.
More recently, MentraSuite~\cite{xiaoMentraSuitePostTrainingLarge2025b} advances mental health reasoning through its MentraBench component, evaluating five core reasoning dimensions including appraisal, diagnosis, intervention, abstraction, and verification, yet primarily assesses reasoning chains rather than end-to-end consultation workflows.
Despite these efforts, critical gaps remain.
Existing benchmarks rarely incorporate clinically used diagnostic labels, provide limited support for multi-turn diagnostic consultation, and seldom adopt agent-based paradigms that decouple patient simulation from diagnostic reasoning---features essential for evaluating LLMs in realistic psychiatric workflows.

\section{Dataset}
\label{sec:dataset}


\subsection{LingxiDiag-Clinical dataset} 

Clinical recordings and reports were collected from approximately 4,500 outpatients at the Shanghai Mental Health Center (SMHC) between 2023 and 2024~\cite{xuIdentifyingPsychiatricManifestations2025c}.
We preprocessed the audio, anonymized personally identifiable information, and transcribed it via automated speech recognition, followed by manual verification to ensure accuracy.
Afterward, we curated 1,709 cases with well-annotated electronic medical records (EMRs) and verified transcriptions to form the LingxiDiag-Clinical dataset.
The study protocol was reviewed and approved by the Ethics Committee of the SMHC Institutional Review Board, ensuring compliance with ethical research standards.
Informed consent was obtained from each participant or their legal guardian prior to participation.

\subsection{LingxiDiag-16K dataset}

To enable scalable evaluation while protecting patient privacy, we generated LingxiDiag-16K, a dataset of 16,000 synthetic consultation dialogues with synthetic EMRs.
The generation process preserved the demographic and clinical distributions observed in the real patient population.
Each case in LingxiDiag-16K includes complete patient profiles comprising demographics, chief complaints, present illness history, past medical and psychiatric history, family history, and diagnostic conclusions.
To align the synthetic distribution with the real data, we first extracted demographic and clinical features from the collected cases to construct a knowledge graph.
We then generate synthetic EMRs by sampling from this graph according to the empirical distribution in the real data (full pipeline in Appendix~\ref{app:synthesis}).
As shown in Table~\ref{tab:demographics}, LingxiDiag-16K closely matches the real data distribution across age groups, gender, and diagnostic groups.
Moreover, LingxiDiag-16K reproduces age-dependent social patterns and linguistic properties of clinical records, as illustrated in Figure~\ref{fig:personal_history} and Figure~\ref{fig:chief_complaint} in the Appendix, respectively.
For LingxiDiag-16K, we generated 16,000 synthetic cases, randomly selected 1,000 samples each for validation and testing, and preserved the same distribution across splits.
Dialogues in LingxiDiag-16K are generated through our multi-agent framework, where a Qwen3-32B-powered American Psychiatric Association (APA)-guided Doctor Agent interacts with the LingxiDiag-Patient, followed by a proper polishing for the inconsistency.
We elaborate on this multi-agent generation framework in the subsequent sections.
To further confirm that LingxiDiag-16K captures clinically realistic patterns beyond surface-level statistics, we provide a cross-dataset transfer validation in Appendix~\ref{app:sft_transfer}.

\begin{table}[htbp]
\centering
\footnotesize
\caption{Demographic and diagnostic comparison between Lingxi-Clinical and LingxiDiag-16K datasets.}
\label{tab:demographics}
\setlength\tabcolsep{2pt}
\begin{tabular*}{\linewidth}{@{\extracolsep{\fill}}llccc@{}}
\toprule
 & Category & LingxiDiag-Clinical & LingxiDiag-16K & Diff \\
\midrule
Sample Size & N & 1,709 & 16,000 & -- \\
\midrule
\multirow{8}{*}{Age} & Mean$\pm$SD & 36.4$\pm$10 & 32.1$\pm$12.0 & -- \\
& 0--18 & 0.8 & 7.7 & +6.8 \\
& 18--25 & 18.3 & 22.4 & +4.1 \\
& 25--35 & 33.7 & 33.6 & $-$0.1 \\
& 35--45 & 24.7 & 19.9 & $-$4.8 \\
& 45--55 & 13.0 & 9.8 & $-$3.3 \\
& 55--65 & 5.9 & 4.4 & $-$1.5 \\
& 65+ & 3.6 & 2.1 & $-$1.5 \\
\midrule
\multirow{2}{*}{Gender} & Male & 32.3 & 32.4 & +0.1 \\
& Female & 67.7 & 67.6 & $-$0.1 \\
\midrule
\multicolumn{2}{l}{F32 (Depressive Episode)} & 35.2 & 34.8 & $-$0.4 \\
\multicolumn{2}{l}{F41 (Anxiety Disorders)} & 22.1 & 22.5 & +0.4 \\
\multicolumn{2}{l}{F43 (Stress-related)} & 12.3 & 12.1 & $-$0.2 \\
\multicolumn{2}{l}{F31 (Bipolar Disorder)} & 8.7 & 8.9 & +0.2 \\
\multicolumn{2}{l}{F42 (OCD)} & 5.4 & 5.6 & +0.2 \\
\multicolumn{2}{l}{F20 (Schizophrenia)} & 4.8 & 4.5 & $-$0.3 \\
\multicolumn{2}{l}{F45 (Somatoform)} & 3.9 & 4.1 & +0.2 \\
\multicolumn{2}{l}{F51 (Sleep Disorders)} & 3.2 & 3.3 & +0.1 \\
\multicolumn{2}{l}{F39 (Unspecified Mood)} & 2.1 & 2.0 & $-$0.1 \\
\multicolumn{2}{l}{F98 (Childhood-onset)} & 1.5 & 1.4 & $-$0.1 \\
\multicolumn{2}{l}{Others} & 0.8 & 0.8 & 0.0 \\
\bottomrule
\end{tabular*}
\end{table}

\subsection{Patient Agent}
\label{sec:patient_agent}

The Patient Agent aims to simulate realistic outpatient behavior during psychiatric consultation.
Following existing work \cite{yinMDD5kNewDiagnostic2025,ferrazMindEvalMultiTurnMental2025}, the patient agent is prompted by a pretrained LLM augmented with structured patient profiles constructed from real clinical data, including demographics, chief complaints, present illness history, and diagnostic information.
In addition to the patient profile, we also provide the conversation history as context to guide response generation, so that the patient agent behaves more like a real patient during the synthetic consultation.
However, compared to real patient responses, we observed that LLM-simulated responses exhibit several unnatural characteristics: (1) they tend to be longer, (2) symptoms are often disclosed all at once rather than gradually, and (3) the language is overly polished and lacks colloquial naturalness.
Therefore, we enhance the patient agent by incorporating a carefully designed set of prompts to elicit more natural responses.
We also control response length by sampling target lengths from the empirical distribution in the real clinical data.
We refer to this enhanced patient agent as LingxiDiag-Patient, which provides a more suitable prompting context than prior approaches~\cite{yinMDD5kNewDiagnostic2025,ferrazMindEvalMultiTurnMental2025}.


\subsection{Doctor Agent}
\label{sec:doctor_agent}

The Doctor Agent simulates a psychiatrist consultation behavior when interacting with the Patient Agent.
We implement four consultation strategies to accommodate different clinical reasoning approaches.
First, the Free-form strategy employs LLMs instructed to act as senior psychiatrists conducting clinical interviews without external guidance.
The Doctor Agent autonomously selects questioning directions based on patient responses and determines when sufficient information has been collected for diagnosis.
Second, we adapt the Symptom-Tree strategy, which uses symptom-based decision trees derived from MDD-5K diagnostic protocols as described in~\cite{yinMDD5kNewDiagnostic2025}.
However, a key limitation of the symptom-tree approach is that it requires a predefined and finite set of symptoms.  
As the number of potential target disorders increases, the clinician may need to query many symptoms to progressively narrow down the diagnostic space.  
To address this limitation, we adopt an APA-guided strategy that follows a five-phase clinical guideline: screening (chief complaints and symptom duration), assessment (core symptom details and functional impairment), deep-dive (specific symptoms and underlying causes), risk assessment (suicide and self-harm screening), and closure (key information confirmation).  
Finally, we evaluate a retrieval-augmented variant of APA-Guided, denoted APA-Guided + MRD-RAG \cite{sun2025multirounddiagnosticragframework}, which provides the retrieved the diagnostic guideline of top 3 potential diagnosis to support the next-question planning during consultation (retrieval procedure detailed in Appendix~\ref{app:retrieval}).
Each phase includes mandatory and optional topics with explicit transition criteria.

\subsection{Diagnosis Agent}
\label{sec:diagnosis_agent}

The Diagnosis Agent performs a psychiatric diagnosis based on complete consultation transcripts between the Doctor and Patient Agents.
Unlike the Doctor Agent, which conducts real-time questioning, the Diagnosis Agent receives the full dialogue history and produces diagnostic conclusions with supporting clinical rationales. 
We use different prompts for the three diagnostic tasks.

\section{Benchmark Evaluation Framework}
\label{sec:benchmark}

\begin{table*}[t]
\centering
\footnotesize
\caption{Patient Agent evaluation results. All dimensions scored 1--5 (higher is better). Overall is the average across all dimensions. Best results are \underline{\textbf{bold with underline}}, second best are \underline{underlined}.}
\label{tab:patient_eval}
\setlength\tabcolsep{2pt}
\begin{tabular*}{0.8\linewidth}{@{\extracolsep{\fill}}llccccccc@{}}
\toprule
Patient Version & Backbone & Accuracy & Honesty & Brevity & Proactivity & Restraint & Polish & Overall \\
\midrule
\multirow{13}{*}{MDD-5K-Patient  \cite{yinMDD5kNewDiagnostic2025}} & Claude-Haiku-4.5 & 4.92$\pm$0.03 & 3.55$\pm$0.22 & 1.18$\pm$0.15 & 1.07$\pm$0.04 & 1.36$\pm$0.17 & 1.33$\pm$0.22 & 2.23 \\
 & Baichuan-M3-235B & 4.78$\pm$0.07 & 3.21$\pm$0.21 & 1.44$\pm$0.11 & 1.34$\pm$0.02 & 1.69$\pm$0.11 & 1.71$\pm$0.23 & 2.36 \\
 & Qwen3-8B & 4.84$\pm$0.11 & 2.91$\pm$0.45 & 1.55$\pm$0.27 & 1.36$\pm$0.05 & 1.67$\pm$0.30 & 1.98$\pm$0.47 & 2.39 \\
 & Baichuan-M2-32B & 4.84$\pm$0.09 & 3.02$\pm$0.03 & 1.55$\pm$0.23 & 1.35$\pm$0.06 & 1.69$\pm$0.19 & 2.02$\pm$0.36 & 2.41 \\
 & Qwen3-1.7B & 4.76$\pm$0.20 & 2.88$\pm$0.21 & 2.14$\pm$0.31 & 1.77$\pm$0.09 & 2.46$\pm$0.45 & 2.54$\pm$0.53 & 2.76 \\
 & Gemini-3-Flash & 4.79$\pm$0.11 & 3.22$\pm$0.35 & 2.44$\pm$0.16 & 2.24$\pm$0.09 & 2.46$\pm$0.18 & 2.60$\pm$0.29 & 2.96 \\
 & Qwen3-32B & 4.63$\pm$0.17 & 2.76$\pm$0.18 & 2.67$\pm$0.31 & 2.46$\pm$0.09 & 2.82$\pm$0.28 & 2.96$\pm$0.31 & 3.05 \\
 & GPT-5-Mini & 4.36$\pm$0.19 & 3.10$\pm$0.45 & 2.77$\pm$0.17 & 2.59$\pm$0.06 & 3.59$\pm$0.26 & 2.83$\pm$0.15 & 3.21 \\
 & Kimi-K2-Thinking & 4.50$\pm$0.20 & 2.97$\pm$0.28 & 3.03$\pm$0.20 & 2.83$\pm$0.09 & 3.13$\pm$0.13 & 3.24$\pm$0.17 & 3.28 \\
 & Qwen3-4B-Think & 4.73$\pm$0.36 & 2.92$\pm$0.94 & 3.08$\pm$0.71 & 2.77$\pm$0.61 & 3.10$\pm$0.61 & 3.17$\pm$0.28 & 3.30 \\
 & GPT-OSS-20B & 4.40$\pm$0.21 & 2.95$\pm$0.21 & 3.28$\pm$0.13 & 3.12$\pm$0.11 & 3.52$\pm$0.26 & 3.29$\pm$0.15 & 3.43 \\
 & Grok-4.1-Fast & 4.53$\pm$0.20 & 3.17$\pm$0.11 & 3.71$\pm$0.18 & 3.46$\pm$0.13 & 3.84$\pm$0.04 & 3.91$\pm$0.10 & 3.77 \\
 & DeepSeek-V3.2 & 4.30$\pm$0.30 & 3.25$\pm$0.02 & 4.22$\pm$0.07 & 4.13$\pm$0.02 & 4.31$\pm$0.05 & 4.31$\pm$0.07 & 4.09 \\
\midrule
\multirow{13}{*}{LingxiDiag-Patient} & Baichuan-M3-235B & 4.90$\pm$0.10 & 4.29$\pm$0.10 & 3.74$\pm$0.14 & 3.67$\pm$0.10 & 3.96$\pm$0.02 & 3.83$\pm$0.10 & 4.07 \\
 & GPT-OSS-20B & 4.77$\pm$0.11 & 4.24$\pm$0.03 & 3.84$\pm$0.11 & 3.77$\pm$0.12 & 4.05$\pm$0.08 & 3.85$\pm$0.12 & 4.09 \\
 & Gemini-3-Flash & 4.37$\pm$0.63 & 4.12$\pm$0.14 & 4.02$\pm$0.18 & 3.92$\pm$0.12 & 4.16$\pm$0.03 & 4.15$\pm$0.09 & 4.12 \\
 & Grok-4.1-Fast & 3.21$\pm$2.27 & 4.09$\pm$0.09 & 4.52$\pm$0.18 & 4.46$\pm$0.17 & 4.65$\pm$0.10 & 4.57$\pm$0.18 & 4.25 \\
 & GPT-5-Mini & \underline{4.94$\pm$0.06} & 4.42$\pm$0.14 & 4.04$\pm$0.18 & 3.87$\pm$0.13 & 4.27$\pm$0.07 & 4.18$\pm$0.13 & 4.29 \\
 & Qwen3-4B & 3.15$\pm$2.24 & 4.11$\pm$0.06 & \underline{4.66$\pm$0.12} & \underline{\textbf{4.63$\pm$0.10}} & \underline{\textbf{4.70$\pm$0.08}} & \underline{\textbf{4.69$\pm$0.13}} & 4.32 \\
 & Claude-Haiku-4.5 & \underline{\textbf{4.95$\pm$0.07}} & 4.43$\pm$0.16 & 4.07$\pm$0.20 & 3.99$\pm$0.14 & 4.30$\pm$0.08 & 4.23$\pm$0.15 & 4.33 \\
 & Kimi-K2-Thinking & 4.90$\pm$0.08 & 4.34$\pm$0.14 & 4.15$\pm$0.17 & 4.11$\pm$0.10 & 4.35$\pm$0.09 & 4.31$\pm$0.10 & 4.36 \\
 & Baichuan-M2-32B & 4.81$\pm$0.17 & 4.26$\pm$0.16 & 4.37$\pm$0.10 & 4.34$\pm$0.09 & 4.45$\pm$0.10 & 4.41$\pm$0.13 & 4.44 \\
 & DeepSeek-V3.2 & 4.92$\pm$0.09 & 4.45$\pm$0.09 & 4.53$\pm$0.16 & 4.50$\pm$0.13 & 4.63$\pm$0.11 & 4.57$\pm$0.15 & 4.60 \\
 & Qwen3-8B & 4.91$\pm$0.10 & \underline{\textbf{4.56$\pm$0.13}} & 4.58$\pm$0.13 & 4.55$\pm$0.09 & 4.61$\pm$0.09 & 4.62$\pm$0.08 & 4.64 \\
 & Qwen3-1.7B & 4.86$\pm$0.09 & 4.50$\pm$0.19 & 4.64$\pm$0.08 & 4.60$\pm$0.07 & 4.67$\pm$0.06 & 4.65$\pm$0.08 & \underline{4.65} \\
 & Qwen3-32B & 4.92$\pm$0.10 & \underline{4.51$\pm$0.19} & \underline{\textbf{4.67$\pm$0.11}} & \underline{4.62$\pm$0.09} & \underline{4.67$\pm$0.09} & \underline{4.66$\pm$0.08} & \underline{\textbf{4.67}} \\
\midrule
\multicolumn{2}{l}{LingxiDiag-Clinical Dataset} & 4.65$\pm$0.20 & 3.93$\pm$0.23 & 4.27$\pm$0.08 & 4.11$\pm$0.09 & 4.51$\pm$0.05 & 4.35$\pm$0.14 & 4.30 \\
\bottomrule
\end{tabular*}
\end{table*}

LingxiDiagBench comprises two evaluation paradigms: static evaluation (LingxiDiagBench-Static) using synthetic EMRs and dialogues as ground truth, and dynamic evaluation (LingxiDiagBench-Dynamic) where models interact with Patient Agents in real-time consultation.
The static evaluation paradigm focuses on reproducible evaluation based on fixed consultation transcripts, while the dynamic evaluation paradigm evaluates end-to-end consultation, where Doctor Agents interact with Patient Agents in real time and then provide diagnostic conclusions.

\subsection{LingxiDiagBench-Static}
\label{sec:static_benchmark}

The LingxiDiagBench-Static benchmark consists of two tasks: assisted diagnosis and doctor next question prediction.
The assisted diagnosis task requires predicting psychiatric diagnoses from complete consultation dialogues, and we evaluate performance at three difficulty levels aligned with clinical diagnostic challenges.
First, the binary classification task distinguishes patients with depression from patients with anxiety, focusing on samples without comorbidity.
This task evaluates the fundamental ability to differentiate between the two most prevalent outpatient psychiatric conditions.
Second, the four-way classification task extends to four categories: pure depression, pure anxiety, mixed depression-anxiety, and other psychiatric conditions.
This task introduces the challenge of recognizing comorbidity patterns and identifying when presentations fall outside the primary diagnostic focus.
Third, the twelve-way classification task spans major ICD-10 psychiatric categories: F20 (schizophrenia), F31 (bipolar disorder), F32 (depressive episode), F39 (unspecified mood disorder), F41 (anxiety disorders), F42 (obsessive-compulsive disorder), F43 (stress-related disorders), F45 (somatoform disorders), F51 (sleep disorders), F98 (childhood-onset disorders), Z71 (counseling), and Others.
This task requires broad differential diagnosis capabilities across heterogeneous conditions, including comorbidity condition prediction. 
We first employ TF-IDF methods to extract features with logistic regression, support vector machines, and random forest classifiers.
For LLMs, we evaluate a broad range of model families, including the Qwen3 series (1.7B, 4B, 8B, 32B)~\cite{yangQwen3TechnicalReport2025}, the Baichuan series (Baichuan-M2-32B, Baichuan-M3-235B)~\cite{teamBaichuanM2ScalingMedical2025}, the Kimi series (K2-Thinking)~\cite{teamKimiK2Open2025}, the DeepSeek series (DeepSeek-V3.2)~\cite{deepseek-aiDeepSeekV32PushingFrontier2025}, Google Gemini (Gemini-3-Flash)~\cite{teamGeminiFamilyHighly2025}, OpenAI GPT (GPT-OSS-20B, GPT-5-Mini)~\cite{openaiGPT4TechnicalReport2024}, and Anthropic Claude (Claude-Haiku-4.5)~\cite{anthropic_claude45_haiku}.
We apply zero-shot inference to evaluate all LLMs on the same 1000 samples in LingxiDiagBench-Static, ensuring a consistent evaluation setting.
We report accuracy, macro F1-score, and weighted F1-score for all classification tasks.
For the twelve-way task, we additionally report Top-1 and Top-3 accuracy to capture cases where the correct diagnosis appears among the top differential diagnoses.

The doctor's next question prediction task evaluates understanding of consultation flow by predicting the next appropriate doctor question given dialogue context. 
This task assesses whether models can generate clinically appropriate follow-up questions that advance diagnostic reasoning.
For the question prediction task, where we test on the same LLMs, we report BLEU, Rouge-L, and BertScore, which measure n-gram overlap, longest common subsequence, and semantic similarity, respectively.

\subsection{LingxiDiagBench-Dynamic}
\label{sec:dynamic_benchmark}

The dynamic benchmark evaluates the performance of Patient Agents, the consultation capabilities of Doctor Agents, and the diagnostic accuracy of Doctor Agents.

\textbf{Patient Agent Evaluation:}
We evaluate Patient Agent quality along six dimensions using an LLM-as-a-Judge protocol.  
To improve robustness, we aggregate judgments from three evaluator models (Gemma-3-27B, GPT-OSS-20B, and Qwen3-30B-A3B)\XSH{, where the aggregation uses a 3-model ensemble with median imputation for missing scores followed by arithmetic averaging across models, applied consistently to both Patient and Doctor Agent evaluation.}
Moreover, based on LingxiDiag-16K, we standardize the prompting context so that different patient agents respond to the same set of doctor questions under identical conditions.  
The evaluation dimensions are organized into two groups.  
Factual consistency includes accuracy and honesty, while the remaining four dimensions---response brevity, information proactivity, emotional restraint, and language polish---evaluate the naturalness of the patient agent responses.

\textbf{Doctor Agent Evaluation: }
Similar to \cite{ferrazMindEvalMultiTurnMental2025}, we evaluate Doctor Agent consultation quality using LLM-as-a-Judge methodology across five clinically relevant dimensions (information completeness, symptom exploration depth, differential diagnosis awareness, risk assessment, and communication quality), each scored on a 1--6 Likert scale \XSH{using above mentioned three evaluator models} as the evaluator.
For dynamic diagnosis, we report the same diagnostic accuracy metrics as static evaluation across three diagnostic tasks: accuracy, macro F1-score, and weighted F1-score for 2 and 4-class classification tasks, as well as accuracy, Top1 accuracy, Top3 accuracy, macro F1-score, and weighted F1-score for 12-class classification tasks (formal metric definitions in Appendix~\ref{app:metrics}).
The key difference is that diagnosis follows interactive consultation rather than analysis of pre-existing transcripts, where the base models not only do diagnosis but also lead the diagnostic consultation.

\section{Results}
\label{sec:results}

\subsection{Patient Agent Evaluation}
\label{sec:patient_results}

In Table~\ref{tab:patient_eval}, we present a comprehensive evaluation of Patient Agent quality across different backbone models and data sources.
All dimensions are scored on a 1--5 scale, with evaluation conducted using a three-model fusion (Gemma-3-27B, GPT-OSS-20B, Qwen3-30B-A3B).
LingxiDiag-Patient Agents substantially outperform their MDD-5K counterparts across all behavioral authenticity dimensions, attributable to refined prompt engineering and output length control, where Qwen3-32B attains the highest Overall score of 4.67.
Most of the LLMs excel in the Accuracy dimension, reflecting superior adherence to patient background profiles.
Real clinical data serves as the ground-truth reference with an Overall score of 4.30, which may be due to occasional misalignment between recorded conversations and EMR histories in authentic clinical settings.
Overall, current LLMs adhere closely to the provided profiles and can appropriately refuse to answer questions when the required information is not available in the background context.
Therefore, in the dynamic benchmark, we utilize Qwen3-32B as the backbone model for the patient agent.


\subsection{Static Benchmark Results}
\label{sec:static_results}

\begin{table*}[t]
\centering
\footnotesize
\caption{LingxiDiagBench--Static evaluation results on LingxiDiag-16K. Best results are \underline{\textbf{bold with underline}}, second best are \underline{underlined}.}
\label{tab:static_synthetic}
\setlength\tabcolsep{2pt}
\begin{tabular*}{0.8\linewidth}{@{\extracolsep{\fill}}lcccccccccccc@{}}
\toprule
& \multicolumn{3}{c}{2-class} & \multicolumn{3}{c}{4-class} & \multicolumn{4}{c}{12-class} & \\
\cmidrule(lr){2-4} \cmidrule(lr){5-7} \cmidrule(lr){8-12}
Method & Acc & m-F1 & w-F1 & Acc & m-F1 & w-F1 & Acc & Top1-Acc & Top3-Acc & m-F1 & w-F1 & Overall \\
\midrule
TF-IDF + SVM & 0.740 & 0.687 & 0.741 & 0.451 & \underline{0.426} & \underline{0.450} & 0.308 & 0.481 & 0.566 & \underline{0.242} & \underline{0.482} & 0.507 \\
TF-IDF + RF & 0.751 & 0.651 & 0.729 & \underline{\textbf{0.479}} & 0.391 & 0.437 & 0.315 & 0.377 & 0.403 & 0.122 & 0.382 & 0.458 \\
TF-IDF + LR & 0.753 & 0.713 & 0.758 & \underline{0.476} & \underline{\textbf{0.458}} & \underline{\textbf{0.480}} & 0.268 & \underline{\textbf{0.496}} & \underline{\textbf{0.645}} & \underline{\textbf{0.295}} & \underline{\textbf{0.520}} & \underline{\textbf{0.533}} \\
Qwen3-1.7B & 0.786 & 0.694 & 0.765 & 0.392 & 0.285 & 0.302 & 0.145 & 0.460 & 0.545 & 0.162 & 0.394 & 0.448 \\
Baichuan-M2-32B & 0.803 & 0.748 & 0.798 & 0.406 & 0.342 & 0.361 & 0.232 & 0.376 & 0.489 & 0.136 & 0.378 & 0.461 \\
Baichuan-M3-235B & 0.816 & 0.764 & 0.811 & 0.390 & 0.371 & 0.387 & 0.254 & 0.393 & 0.514 & 0.143 & 0.396 & 0.476 \\
Qwen3-4B & 0.825 & 0.766 & 0.815 & 0.401 & 0.332 & 0.352 & 0.021 & 0.475 & \underline{0.637} & 0.168 & 0.422 & 0.474 \\
Qwen3-8B & 0.835 & 0.776 & 0.824 & 0.408 & 0.335 & 0.355 & 0.012 & 0.459 & 0.599 & 0.177 & 0.420 & 0.473 \\
GPT-OSS-20B & 0.778 & 0.747 & 0.784 & 0.402 & 0.355 & 0.367 & 0.259 & 0.463 & 0.523 & 0.181 & 0.408 & 0.479 \\
Kimi-K2-Think & 0.818 & 0.760 & 0.810 & 0.409 & 0.374 & 0.391 & 0.335 & 0.427 & 0.468 & 0.155 & 0.379 & 0.484 \\
DeepSeek-V3.2 & 0.820 & 0.788 & 0.823 & 0.441 & 0.400 & 0.412 & 0.323 & 0.438 & 0.489 & 0.164 & 0.408 & 0.501 \\
GPT-5-Mini & 0.803 & 0.747 & 0.797 & 0.434 & 0.372 & 0.387 & \underline{\textbf{0.409}} & 0.487 & 0.505 & 0.188 & 0.418 & 0.504 \\
Gemini-3-Flash & \underline{\textbf{0.854}} & \underline{\textbf{0.816}} & \underline{\textbf{0.851}} & 0.422 & 0.390 & 0.407 & 0.172 & \underline{0.492} & 0.574 & 0.197 & 0.439 & 0.510 \\
Qwen3-32B & 0.827 & 0.791 & 0.827 & 0.438 & 0.384 & 0.404 & 0.241 & 0.470 & 0.566 & 0.188 & 0.431 & 0.506 \\
Claude-Haiku-4.5 & 0.825 & 0.783 & 0.823 & 0.444 & 0.401 & 0.417 & \underline{0.395} & 0.478 & 0.501 & 0.199 & 0.412 & 0.516 \\
Grok-4.1-Fast & \underline{0.841} & \underline{0.799} & \underline{0.838} & 0.470 & 0.424 & 0.439 & 0.351 & 0.465 & 0.495 & 0.195 & 0.409 & \underline{0.521} \\
\bottomrule
\end{tabular*}
\end{table*}

\begin{table*}[t]
\centering
\footnotesize
\caption{LingxiDiagBench--Static evaluation results on LingxiDiag-Clinical dataset. Best results are \underline{\textbf{bold with underline}}, second best are \underline{underlined}.}
\label{tab:static_real}
\setlength\tabcolsep{2pt}
\begin{tabular*}{\linewidth}{@{\extracolsep{\fill}}lcccccccccccc@{}}
\toprule
& \multicolumn{3}{c}{2-class} & \multicolumn{3}{c}{4-class} & \multicolumn{4}{c}{12-class} & \\
\cmidrule(lr){2-4} \cmidrule(lr){5-7} \cmidrule(lr){8-12}
Method & Acc & m-F1 & w-F1 & Acc & m-F1 & w-F1 & Acc & Top1-Acc & Top3-Acc & m-F1 & w-F1 & Overall \\
\midrule
TF-IDF + LR & 0.724 & 0.663 & 0.735 & 0.426 & 0.422 & 0.426 & \underline{0.299} & 0.442 & 0.519 & \underline{0.275} & 0.469 & 0.491 \\
TF-IDF + SVM & 0.787 & 0.724 & 0.790 & 0.424 & 0.407 & 0.422 & \underline{\textbf{0.320}} & 0.413 & 0.447 & 0.266 & 0.436 & 0.494 \\
TF-IDF + RF & 0.769 & 0.551 & 0.708 & 0.417 & 0.314 & 0.360 & 0.231 & 0.261 & 0.270 & 0.227 & 0.323 & 0.403 \\
GPT-OSS-20B & 0.824 & 0.785 & 0.831 & 0.254 & 0.102 & 0.103 & 0.050 & 0.050 & 0.052 & 0.009 & 0.008 & 0.279 \\
Baichuan-M2-32B & 0.824 & 0.785 & 0.831 & \underline{0.494} & \underline{0.476} & \underline{0.492} & 0.247 & 0.431 & 0.576 & 0.169 & 0.443 & 0.524 \\
Qwen3-8B & 0.851 & 0.808 & 0.854 & 0.456 & 0.451 & 0.446 & 0.041 & \underline{0.512} & \underline{0.669} & 0.256 & \underline{0.474} & 0.529 \\
Qwen3-1.7B & \underline{0.882} & \underline{\textbf{0.846}} & \underline{0.884} & 0.458 & 0.420 & 0.422 & 0.166 & 0.506 & 0.610 & 0.204 & 0.456 & 0.532 \\
Qwen3-4B & \underline{\textbf{0.887}} & \underline{0.842} & \underline{\textbf{0.884}} & 0.458 & 0.456 & 0.454 & 0.063 & \underline{\textbf{0.522}} & \underline{\textbf{0.698}} & 0.244 & \underline{\textbf{0.478}} & \underline{0.544} \\
Qwen3-32B & 0.824 & 0.780 & 0.829 & \underline{\textbf{0.524}} & \underline{\textbf{0.526}} & \underline{\textbf{0.523}} & 0.204 & 0.472 & 0.601 & \underline{\textbf{0.278}} & 0.470 & \underline{\textbf{0.548}} \\
\bottomrule
\end{tabular*}
\end{table*}

We evaluate AI-assisted diagnosis on both LingxiDiag-16K and the LingxiDiag-Clinical dataset using both traditional frequency-based methods and LLMs (shown in Table~\ref{tab:static_synthetic} and Table~\ref{tab:static_real}, respectively).
For differential depression and anxiety, accuracy is high across methods, reaching 0.854 on LingxiDiag-16K (Gemini-3-Flash) and 0.887 on LingxiDiag-Clinical (Qwen3-4B).
For the 4-class task, performance drops to around 0.39--0.48 accuracy, with TF-IDF + RF achieving 0.479 on LingxiDiag-16K and Qwen3-32B achieving 0.524 on LingxiDiag-Clinical.
For the 12-class task, the best 12-class accuracy on LingxiDiag-16K is 0.409 (GPT-5-Mini), while the best macro F1 on LingxiDiag-Clinical is 0.278 (Qwen3-32B), and the best Top-3 accuracy is 0.698 (Qwen3-4B).
Overall, TF-IDF + LR attains the best Overall score on LingxiDiag-16K (0.533), and Qwen3-32B achieves the best Overall score on LingxiDiag-Clinical (0.548).

Table~\ref{tab:question_pred} presents results for the psychiatrist's follow-up question prediction, which evaluates models' understanding of consultation flow on both synthetic (LingxiDiag-16K) and real clinical (LingxiDiag-Clinical) data.
Overall, models demonstrate comparable performance, with BLEU scores ranging from approximately 20\% to 23\% and BertScore ranging from 72\% to 84\%.

\begin{table*}[t]
\centering
\footnotesize
\caption{Doctor question prediction task results on LingxiDiag-16K and LingxiDiag-Clinical Dataset. Models not evaluated on LingxiDiag-Clinical Dataset are marked with (---). Best results are \underline{\textbf{bold with underline}}, second best are \underline{underlined}.}
\label{tab:question_pred}
\setlength\tabcolsep{3pt}
\begin{tabular*}{0.8\linewidth}{@{\extracolsep{\fill}}lcccccc@{}}
\toprule
& \multicolumn{3}{c}{LingxiDiag-16K} & \multicolumn{3}{c}{LingxiDiag-Clinical Dataset} \\
\cmidrule(lr){2-4} \cmidrule(lr){5-7}
Model & BLEU (\%) & Rouge-L (\%) & BertScore (\%) & BLEU (\%) & Rouge-L (\%) & BertScore (\%) \\
\midrule
GPT-OSS-20B & 19.7 & 15.3 & 72.2 & 33.1 & 10.2 & 71.5 \\
Baichuan-M2-32B & 20.5 & 21.0 & 81.8 & 28.7 & 12.3 & 72.6 \\
Baichuan-M3-235B & 20.6 & 20.3 & 81.0 & --& --& --\\
Gemini-3-Flash & 21.3 & 20.0 & 81.9 & --& --& --\\
GPT-5-Mini & 21.5 & 18.1 & 79.7 & --& --& --\\
Grok-4.1-Fast & 21.6 & 22.5 & 82.6 & --& --& --\\
Qwen3-4B & 21.6 & 23.5 & 84.2 & \underline{33.4} & \underline{13.5} & \underline{\textbf{77.6}} \\
Qwen3-8B & 21.7 & \underline{\textbf{24.9}} & \underline{\textbf{84.4}} & 32.3 & \underline{\textbf{14.0}} & \underline{77.3} \\
Qwen3-32B & 21.7 & 23.0 & 83.5 & 32.1 & 13.0 & 76.3 \\
Kimi-K2-Thinking & 21.8 & 22.3 & 83.3 & --& --& --\\
Qwen3-1.7B & 21.9 & 22.2 & 83.2 & \underline{\textbf{33.5}} & 13.0 & 76.3 \\
DeepSeek-V3.2 & \underline{22.2} & \underline{24.6} & \underline{84.2} & --& --& --\\
Claude-Haiku-4.5 & \underline{\textbf{22.7}} & 21.0 & 82.9 & --& --& --\\
\bottomrule
\end{tabular*}
\end{table*}


\subsection{Dynamic Benchmark Results}
\label{sec:dynamic_results}

Table~\ref{tab:dynamic_results} presents the comprehensive evaluation of Doctor Agents across both LLM-as-a-Judge consultation quality dimensions and complete diagnostic classification performance, including all metrics for 2-class, 4-class, and 12-class tasks.
We compare four doctor strategies, including Free-form, Symptom-Tree, APA-Guided, and APA-Guided + MRD-RAG.

\begin{table*}[h]
\centering
\caption{Dynamic benchmark results. LLM-as-a-Judge dimensions scored 1--6 (higher is better). Classification metrics reported as percentage (\%). Best results are \underline{\textbf{bold with underline}}, second best are \underline{underlined}. Results sorted by Clf-Ovl within each strategy.}
\label{tab:dynamic_results}
\setlength\tabcolsep{1pt}
\footnotesize
\begin{tabular*}{\linewidth}{@{\extracolsep{\fill}}llccccccccccccccccccc@{}}
\toprule
& & \multicolumn{6}{c}{LLM-as-a-Judge (1--6)} & \multicolumn{3}{c}{2-class (\%)} & \multicolumn{3}{c}{4-class (\%)} & \multicolumn{5}{c}{12-class (\%)} & \\
\cmidrule(lr){3-8} \cmidrule(lr){9-11} \cmidrule(lr){12-14} \cmidrule(lr){15-19}
Strategy & Model & Clin & Eth & Ass & All & Com & LLM-Ovl & Acc & m-F1 & w-F1 & Acc & m-F1 & w-F1 & Acc & Top1-Acc & Top3-Acc & m-F1 & w-F1 & Clf-Ovl \\
\midrule
\multirow{13}{*}{Free-form} & Qwen3-8B & 2.97 & 4.88 & 2.83 & 2.37 & 2.21 & 3.05 & 78.8 & 57.2 & 73.6 & 24.5 & 20.4 & 16.0 & 1.5 & 22.5 & 33.5 & 10.7 & 14.9 & 28.4 \\
 & Qwen3-4B & 2.96 & 4.81 & 2.72 & 2.40 & 2.18 & 3.01 & 78.8 & 57.2 & 73.6 & 19.0 & 17.1 & 13.8 & 2.5 & 24.0 & 39.5 & 15.9 & 18.0 & 29.0 \\
 & Qwen3-1.7B & 2.70 & 4.43 & 2.00 & 1.87 & 1.85 & 2.57 & 80.8 & 69.0 & 79.3 & 17.5 & 17.3 & 10.1 & 14.0 & 24.0 & 33.0 & 11.5 & 15.8 & 29.6 \\
 & Baichuan-M2-32B & 3.06 & 4.46 & 3.30 & 2.64 & 2.84 & 3.26 & 80.8 & 63.1 & 76.8 & 29.0 & 24.4 & 25.5 & 12.5 & 23.5 & 31.0 & 13.2 & 16.7 & 32.3 \\
 & GPT-OSS-20B & 3.12 & 4.78 & 3.33 & 2.73 & 2.64 & 3.32 & 80.8 & 66.4 & 78.2 & 22.0 & 19.9 & 16.7 & 10.0 & 26.5 & 37.5 & 19.7 & 21.1 & 32.4 \\
 & Kimi-K2-Thinking & 3.69 & 4.88 & 3.85 & 3.04 & \underline{3.14} & 3.72 & 84.1 & 71.9 & 81.1 & \underline{36.0} & 30.7 & \underline{36.5} & 18.0 & 19.0 & 20.5 & 12.7 & 13.9 & 34.5 \\
 & Claude-Haiku-4.5 & 3.53 & 4.96 & 3.80 & 3.03 & 2.92 & 3.65 & 84.1 & 76.5 & 83.2 & 27.5 & 24.5 & 23.7 & 21.5 & 24.5 & 29.0 & 16.0 & 17.4 & 34.7 \\
 & Gemini-3-Flash & 3.73 & 4.98 & 3.81 & 3.16 & 2.95 & 3.73 & 84.1 & 74.5 & 82.3 & 28.5 & 28.3 & 23.8 & 18.0 & 23.5 & 30.0 & 17.3 & 19.3 & 34.9 \\
 & Baichuan-M3-235B & 3.08 & 4.96 & 2.93 & 2.68 & 2.43 & 3.22 & 80.8 & 63.1 & 76.8 & 33.0 & 29.4 & 35.3 & 19.0 & 24.5 & 28.5 & 16.3 & 18.3 & 35.1 \\
 & GPT-5-Mini & 3.21 & 4.42 & 3.43 & 2.84 & 3.04 & 3.39 & 86.4 & 79.0 & 85.2 & 27.5 & 20.8 & 22.8 & 25.0 & 26.5 & 28.0 & 19.0 & 18.9 & 35.6 \\
 & Qwen3-32B & 3.13 & 4.91 & 3.19 & 2.63 & 2.40 & 3.25 & 86.5 & 77.5 & 85.2 & 28.0 & 23.5 & 23.7 & 19.5 & 28.5 & 34.0 & 18.3 & 20.6 & 36.2 \\
 & DeepSeek-V3.2 & 3.55 & 5.00 & 3.75 & 3.10 & 2.91 & 3.66 & 84.6 & 70.5 & 81.5 & 35.5 & 31.0 & 32.1 & 23.5 & 30.0 & 35.0 & 21.0 & 22.2 & 38.8 \\
 & Grok-4.1-Fast & 3.13 & 4.72 & 3.57 & 2.92 & 2.97 & 3.46 & 88.6 & 84.4 & 88.5 & 34.0 & 33.4 & 32.4 & 25.5 & 27.5 & 31.5 & 21.1 & 21.1 & 40.1 \\
\midrule
\multirow{13}{*}{Symptom-Tree \cite{yinMDD5kNewDiagnostic2025}} & Qwen3-1.7B & 2.83 & 4.53 & 2.70 & 2.38 & 2.20 & 2.93 & 78.8 & 57.2 & 73.6 & 21.5 & 14.4 & 9.9 & 7.0 & 20.5 & 29.0 & 7.1 & 11.7 & 26.1 \\
 & Qwen3-4B & 2.97 & 4.55 & 3.16 & 2.62 & 2.56 & 3.17 & 76.9 & 65.4 & 76.2 & 20.5 & 19.0 & 13.9 & 0.5 & 23.5 & 38.0 & 14.6 & 16.7 & 29.2 \\
 & GPT-OSS-20B & 3.06 & 4.64 & 3.42 & 2.65 & 3.05 & 3.36 & 80.8 & 66.4 & 78.2 & 25.0 & 22.9 & 18.3 & 12.0 & 25.0 & 31.5 & 13.9 & 16.8 & 31.6 \\
 & Qwen3-8B & 2.97 & 4.47 & 3.21 & 2.47 & 2.64 & 3.15 & 86.5 & 79.1 & 85.8 & 20.5 & 20.2 & 12.8 & 0.5 & 24.5 & 38.0 & 16.3 & 16.9 & 31.7 \\
 & GPT-5-Mini & 3.55 & 4.83 & 3.63 & 2.88 & 2.89 & 3.56 & 81.8 & 72.1 & 80.3 & 26.5 & 23.5 & 21.9 & 20.5 & 23.5 & 25.0 & 16.3 & 16.0 & 32.9 \\
 & Qwen3-32B & 3.23 & 4.88 & 3.66 & 2.97 & 2.75 & 3.50 & 82.7 & 74.9 & 82.4 & 29.0 & 25.0 & 22.1 & 16.0 & 26.5 & 34.0 & 14.5 & 17.2 & 34.4 \\
 & Baichuan-M2-32B & 2.94 & 4.06 & 2.95 & 2.54 & 2.58 & 3.01 & 80.8 & 69.0 & 79.3 & 30.5 & 29.1 & 25.4 & 16.5 & 26.0 & 31.0 & 15.4 & 18.8 & 34.5 \\
 & Kimi-K2-Thinking & 3.55 & 4.64 & 3.62 & 3.07 & 3.08 & 3.59 & 84.1 & 78.1 & 83.8 & 29.5 & 26.1 & 28.8 & 21.5 & 22.5 & 24.0 & 15.3 & 17.2 & 34.9 \\
 & Gemini-3-Flash & 3.57 & 4.75 & 3.75 & 3.03 & 2.88 & 3.60 & 86.4 & 79.0 & 85.2 & 28.0 & 30.1 & 23.0 & 15.0 & 24.0 & 33.5 & 17.0 & 19.5 & 35.7 \\
 & Baichuan-M3-235B & 3.16 & 4.77 & 3.64 & 2.89 & 3.02 & 3.50 & 86.5 & 75.4 & 84.3 & 30.5 & 27.6 & 32.5 & 20.0 & 25.5 & 33.0 & 14.1 & 19.1 & 36.6 \\
 & Claude-Haiku-4.5 & 2.99 & 4.47 & 3.21 & 2.51 & 3.00 & 3.24 & 86.4 & 79.0 & 85.2 & 30.0 & 26.2 & 27.2 & 25.0 & 29.0 & 30.0 & 19.0 & 19.0 & 37.2 \\
 & Grok-4.1-Fast & 2.78 & 4.16 & 3.04 & 2.20 & 2.72 & 2.98 & 88.6 & 83.2 & 88.0 & 30.0 & 29.5 & 27.1 & 26.0 & 28.0 & 28.0 & 18.0 & 19.2 & 37.9 \\
 & DeepSeek-V3.2 & 3.28 & 4.76 & 3.71 & 2.90 & 3.06 & 3.54 & 86.5 & 80.5 & 86.3 & 31.0 & 31.2 & 25.5 & 21.5 & 29.0 & 34.5 & 17.0 & 21.6 & 38.0 \\
\midrule
\multirow{13}{*}{APA-Guided} & Kimi-K2-Thinking & 3.82 & 4.92 & 3.91 & 3.24 & 3.14 & 3.81 & 77.3 & 65.1 & 75.4 & 25.0 & 19.9 & 26.3 & 15.5 & 18.0 & 18.5 & 13.2 & 13.6 & 29.5 \\
 & Qwen3-1.7B & 2.20 & 3.95 & 1.76 & 1.76 & 1.90 & 2.31 & 84.6 & 73.1 & 82.6 & 25.0 & 19.8 & 17.4 & 7.0 & 23.0 & 32.0 & 9.4 & 15.6 & 30.9 \\
 & Qwen3-8B & 3.02 & 4.50 & 3.20 & 2.62 & 2.40 & 3.15 & 82.7 & 76.3 & 82.9 & 20.0 & 19.8 & 16.4 & 0.5 & 28.0 & 40.0 & 15.2 & 18.3 & 31.9 \\
 & Qwen3-4B & 3.06 & 4.86 & 3.37 & 2.80 & 2.55 & 3.33 & 82.7 & 74.9 & 82.4 & 20.5 & 23.1 & 15.5 & 0.0 & 25.0 & 39.5 & 15.5 & 18.0 & 34.2 \\
 & GPT-OSS-20B & 3.36 & 4.91 & 3.73 & 2.84 & 2.92 & 3.55 & 80.8 & 76.6 & 81.9 & 21.5 & 23.6 & 20.0 & 10.0 & 32.0 & 38.5 & 17.7 & 20.0 & 34.3 \\
 & Gemini-3-Flash & 3.91 & 4.99 & 4.05 & 3.33 & 3.13 & 3.88 & 81.8 & 72.1 & 80.3 & 26.5 & 28.1 & 22.9 & 14.5 & 26.0 & 33.0 & 21.2 & 22.0 & 35.0 \\
 & Baichuan-M2-32B & 3.25 & 4.56 & 3.48 & 2.78 & 2.99 & 3.41 & 82.7 & 68.4 & 79.8 & 31.0 & 28.4 & 30.3 & 12.0 & 24.5 & 36.0 & 16.9 & 19.8 & 35.3 \\
 & GPT-5-Mini & 3.85 & 4.96 & 3.90 & 2.97 & 2.97 & 3.73 & 75.0 & 65.6 & 74.6 & 30.0 & 25.6 & 30.1 & 23.5 & 27.5 & 30.5 & 23.3 & 22.3 & 35.6 \\
 & Claude-Haiku-4.5 & 3.67 & 4.93 & 3.97 & 3.03 & 3.01 & 3.72 & 81.8 & 75.8 & 81.8 & 27.0 & 24.6 & 26.8 & 23.5 & 29.0 & 33.5 & 19.4 & 20.2 & 36.4 \\
 & Grok-4.1-Fast & 3.38 & 5.00 & 3.96 & 3.10 & 2.99 & 3.69 & 81.8 & 75.8 & 81.8 & 30.0 & 27.6 & 29.4 & 24.0 & 29.0 & 32.0 & \underline{25.0} & 22.7 & 37.9 \\
 & Baichuan-M3-235B & 3.69 & \underline{5.01} & 3.99 & 3.20 & 3.01 & 3.78 & 88.5 & 81.4 & 87.6 & 34.5 & 29.5 & 34.3 & 18.5 & 27.0 & 32.0 & 16.0 & 18.9 & 38.2 \\
 & Qwen3-32B & 3.52 & 4.91 & 3.92 & 3.12 & 2.96 & 3.69 & 78.8 & 74.7 & 80.2 & 31.5 & 30.4 & 31.6 & 17.5 & 34.0 & \underline{\textbf{44.0}} & 21.9 & 24.4 & 39.1 \\
 & DeepSeek-V3.2 & 3.70 & 4.99 & 4.06 & 3.27 & 2.99 & 3.80 & 88.5 & 85.3 & 89.0 & 31.5 & 29.9 & 29.4 & 23.0 & 32.0 & 38.5 & \underline{\textbf{25.6}} & \underline{\textbf{26.5}} & 41.2 \\
\midrule
\multirow{13}{*}{\makecell{APA-Guided \\+ MRD-RAG \cite{sun2025multirounddiagnosticragframework}}} & Qwen3-8B & 2.78 & 4.39 & 2.20 & 2.11 & 1.75 & 2.65 & 73.1 & 64.1 & 73.8 & 24.0 & 23.4 & 20.7 & 0.5 & 24.5 & 37.0 & 14.7 & 17.4 & 30.3 \\
 & Qwen3-1.7B & 2.27 & 4.13 & 1.82 & 1.69 & 1.95 & 2.37 & 78.8 & 69.3 & 78.5 & 24.5 & 20.5 & 15.7 & 16.0 & 27.0 & 33.0 & 10.2 & 16.3 & 31.4 \\
 & Qwen3-4B & 3.00 & 4.86 & 2.70 & 2.70 & 1.99 & 3.05 & 82.7 & 74.9 & 82.4 & 23.0 & 23.8 & 19.8 & 1.0 & 30.0 & 40.5 & 16.9 & 18.5 & 33.4 \\
 & GPT-OSS-20B & 3.48 & 4.93 & 3.82 & 2.87 & 2.95 & 3.61 & 79.0 & 74.0 & 80.0 & 26.0 & 26.0 & 23.0 & 12.0 & 31.0 & 39.0 & 19.0 & 22.0 & 35.3 \\
 & Baichuan-M2-32B & 3.34 & 4.63 & 3.52 & 2.85 & 3.01 & 3.47 & 80.8 & 69.0 & 79.3 & 31.0 & 29.9 & 29.0 & 12.5 & 25.5 & 39.5 & 17.0 & 21.4 & 35.9 \\
 & GPT-5-Mini & \underline{\textbf{3.98}} & 5.00 & 4.06 & 3.08 & 3.06 & 3.84 & 80.8 & 71.2 & 80.1 & 32.5 & 30.1 & 30.1 & 23.0 & 31.0 & 34.5 & 19.2 & 22.2 & 37.7 \\
 & Baichuan-M3-235B & 3.67 & 4.99 & 3.94 & 3.18 & 3.01 & 3.76 & 86.5 & 79.1 & 85.8 & 31.5 & 29.0 & 33.2 & 19.5 & 29.0 & 34.0 & 17.7 & 21.9 & 38.3 \\
 & Kimi-K2-Thinking & 3.90 & 4.96 & 4.05 & 3.30 & \underline{\textbf{3.21}} & 3.88 & \underline{90.4} & 85.1 & 89.9 & 32.0 & 31.8 & 29.2 & 24.5 & 29.5 & 31.0 & 16.6 & 21.1 & 39.3 \\
 & Gemini-3-Flash & \underline{3.95} & 5.01 & \underline{\textbf{4.18}} & \underline{3.38} & 3.04 & \underline{\textbf{3.91}} & 88.5 & 82.7 & 88.1 & 31.0 & 31.0 & 30.8 & 17.0 & 32.5 & 41.0 & 22.1 & 24.1 & 40.2 \\
 & Qwen3-32B & 3.44 & 4.97 & 3.84 & 3.12 & 2.87 & 3.65 & 82.7 & 78.5 & 83.6 & 35.5 & \underline{35.6} & 32.5 & 21.0 & 32.5 & \underline{43.0} & 21.7 & 24.4 & 40.9 \\
 & DeepSeek-V3.2 & 3.86 & 4.99 & \underline{4.12} & \underline{\textbf{3.39}} & 3.05 & \underline{3.88} & \underline{\textbf{92.3}} & \underline{\textbf{89.7}} & \underline{\textbf{92.5}} & 34.5 & 32.9 & 33.3 & 23.0 & 32.0 & 36.0 & 24.7 & 24.8 & 42.4 \\
 & Claude-Haiku-4.5 & 3.87 & 5.00 & 4.06 & 3.12 & 3.02 & 3.81 & 90.4 & \underline{87.5} & \underline{90.7} & 35.5 & 35.3 & 34.4 & \underline{28.0} & \underline{34.5} & 36.5 & 19.8 & 23.8 & \underline{42.7} \\
 & Grok-4.1-Fast & 3.75 & \underline{\textbf{5.01}} & 4.06 & 3.19 & 3.02 & 3.81 & 88.5 & 83.8 & 88.5 & \underline{\textbf{43.0}} & \underline{\textbf{42.0}} & \underline{\textbf{40.7}} & \underline{\textbf{28.5}} & \underline{\textbf{37.5}} & 40.5 & 22.0 & \underline{25.5} & \underline{\textbf{45.4}} \\
\bottomrule
\end{tabular*}
\end{table*}

\begin{table*}[htbp]
\centering
\caption{LingxiDiagBench--Dynamic evaluation results on LingxiDiag-Clinical dataset with real EMRs and dialogues role-playing.
LLM-as-a-Judge dimensions scored 1--6 (higher is better).
Classification metrics reported as percentage (\%).
Best results are \underline{\textbf{bold with underline}}, second best are \underline{underlined}.
Results sorted by Clf-Ovl within each strategy.}
\label{tab:dynamic_real}
\setlength\tabcolsep{1pt}
\footnotesize
\begin{tabular*}{\linewidth}{@{\extracolsep{\fill}}llccccccccccccccccccc@{}}
\toprule
& & \multicolumn{6}{c}{LLM-as-a-Judge (1--6)} & \multicolumn{3}{c}{2-class (\%)} & \multicolumn{3}{c}{4-class (\%)} & \multicolumn{5}{c}{12-class (\%)} & \\
\cmidrule(lr){3-8} \cmidrule(lr){9-11} \cmidrule(lr){12-14} \cmidrule(lr){15-19}
Strategy & Model & Clin & Eth & Ass & All & Com & LLM-Ovl & Acc & m-F1 & w-F1 & Acc & m-F1 & w-F1 & Acc & Top1-Acc & Top3-Acc & m-F1 & w-F1 & Clf-Ovl \\
\midrule
\multirow{7}{*}{Free-form} & Qwen3-1.7B & 2.26 & 3.42 & 1.65 & 1.52 & 1.53 & 2.08 & 71.2 & 66.9 & 72.9 & 36.5 & 35.5 & 34.1 & 38.5 & 38.5 & 46.0 & 15.4 & 30.4 & 41.6 \\
 & Qwen3-4B & 3.57 & 4.75 & 3.62 & 2.99 & \underline{3.89} & 3.76 & 77.5 & 70.0 & 77.5 & 36.5 & 33.0 & 32.9 & 40.5 & 40.5 & 50.0 & 17.6 & 31.3 & 43.2 \\
 & Baichuan-M3-235B & 3.46 & 4.65 & 3.06 & 2.62 & 2.54 & 3.27 & 85.0 & 76.6 & 83.6 & \underline{43.5} & 38.8 & \underline{43.1} & 34.5 & 34.5 & 46.0 & 16.3 & 28.0 & 44.8 \\
 & GPT-OSS-20B & 3.67 & 4.82 & 3.61 & 2.90 & 3.06 & 3.61 & 86.2 & 78.0 & 84.7 & 39.0 & 38.4 & 37.3 & 43.0 & 43.0 & 54.0 & 18.5 & 32.5 & 47.2 \\
 & Qwen3-32B & 3.78 & \underline{4.94} & \underline{4.13} & \underline{3.45} & \underline{\textbf{4.08}} & \underline{4.08} & 80.0 & 71.3 & 79.2 & 42.5 & 36.6 & 40.0 & 44.5 & 44.5 & 51.0 & 25.5 & 35.4 & 47.4 \\
 & Baichuan-M2-32B & 3.63 & 4.72 & 3.65 & 3.04 & 3.34 & 3.68 & 87.5 & 81.3 & 86.7 & 41.0 & 37.0 & 39.9 & 44.0 & 44.0 & 53.0 & 26.0 & 34.1 & 48.9 \\
 & Qwen3-8B & 3.63 & 4.76 & 3.71 & 3.00 & 3.70 & 3.76 & \underline{88.8} & \underline{84.7} & \underline{88.7} & 40.0 & 37.5 & 36.4 & 43.0 & 43.0 & 54.5 & \underline{26.9} & 33.7 & \underline{49.0} \\
\midrule
\multirow{6}{*}{Symptom-Tree} & Qwen3-1.7B & 2.87 & 3.91 & 2.39 & 2.08 & 1.95 & 2.64 & 80.0 & 68.8 & 78.1 & 31.5 & 25.6 & 25.6 & 36.0 & 36.0 & 45.0 & 9.4 & 23.3 & 38.4 \\
 & Qwen3-4B & 3.24 & 4.25 & 3.03 & 2.53 & 2.67 & 3.14 & 81.2 & 71.4 & 79.8 & 34.5 & 33.0 & 31.8 & 39.5 & 39.5 & 55.5 & 13.0 & 27.1 & 42.9 \\
 & Qwen3-8B & 3.45 & 4.47 & 3.34 & 2.80 & 2.91 & 3.39 & 82.5 & 74.9 & 81.8 & 33.5 & 32.1 & 30.4 & 41.5 & 41.5 & 52.0 & 18.5 & 30.1 & 43.9 \\
 & Qwen3-32B & 3.55 & 4.66 & 3.68 & 3.08 & 3.35 & 3.66 & 82.5 & 75.8 & 82.2 & 36.0 & 35.4 & 34.3 & \underline{\textbf{47.0}} & \underline{\textbf{47.0}} & 56.0 & 24.8 & \underline{37.5} & 47.8 \\
 & Baichuan-M2-32B & 3.20 & 3.81 & 3.08 & 2.72 & 2.75 & 3.11 & \underline{88.8} & \underline{83.5} & \underline{88.2} & \underline{\textbf{44.0}} & \underline{41.9} & 42.4 & 41.0 & 41.0 & 48.0 & 19.5 & 32.8 & 48.4 \\
 & GPT-OSS-20B & 3.57 & 4.68 & 3.58 & 2.91 & 3.49 & 3.65 & \underline{\textbf{91.2}} & \underline{\textbf{87.7}} & \underline{\textbf{91.0}} & 43.0 & 40.8 & 40.3 & 44.5 & 44.5 & 50.0 & 20.8 & 36.2 & \underline{\textbf{50.0}} \\
\midrule
\multirow{6}{*}{APA-Guided} & Qwen3-1.7B & 1.96 & 3.05 & 1.42 & 1.43 & 1.39 & 1.85 & 80.0 & 74.9 & 80.6 & 37.5 & 33.7 & 33.4 & 38.0 & 38.0 & 41.5 & 14.2 & 28.8 & 42.2 \\
 & Qwen3-4B & 3.13 & 4.23 & 3.10 & 2.79 & 2.51 & 3.15 & 78.8 & 73.7 & 79.5 & 32.0 & 31.3 & 29.4 & 41.0 & 41.0 & \underline{\textbf{59.5}} & 17.1 & 30.8 & 43.7 \\
 & Qwen3-8B & 3.24 & 4.00 & 3.05 & 2.65 & 2.39 & 3.07 & 78.8 & 74.4 & 79.7 & 32.5 & 31.7 & 31.0 & 41.5 & 41.5 & 58.0 & 21.3 & 31.9 & 44.5 \\
 & GPT-OSS-20B & 3.79 & 4.82 & 4.07 & 3.43 & 3.57 & 3.94 & 80.0 & 76.2 & 81.0 & 34.5 & 35.1 & 33.8 & 43.5 & 43.5 & 54.5 & 21.7 & 34.4 & 45.9 \\
 & Baichuan-M2-32B & 3.80 & 4.69 & 4.04 & \underline{3.63} & 3.76 & 3.98 & 82.5 & 76.7 & 82.5 & 43.0 & \underline{\textbf{42.2}} & 42.3 & 40.5 & 40.5 & 51.0 & 23.6 & 34.8 & 47.9 \\
 & Qwen3-32B & 3.71 & 4.70 & 4.01 & 3.53 & 3.54 & 3.90 & 80.0 & 76.7 & 81.1 & 36.0 & 35.3 & 35.8 & \underline{46.5} & \underline{46.5} & 54.5 & \underline{\textbf{31.6}} & \underline{\textbf{38.2}} & 48.3 \\
\midrule
\multirow{6}{*}{\makecell{APA-Guided \\+ MRD-RAG}} & Qwen3-8B & 3.22 & 4.09 & 3.09 & 2.68 & 2.43 & 3.10 & 72.5 & 68.7 & 74.1 & 31.0 & 30.7 & 29.5 & 40.0 & 40.0 & 54.5 & 17.4 & 31.2 & 41.8 \\
 & Qwen3-1.7B & 2.30 & 3.34 & 1.69 & 1.74 & 1.48 & 2.11 & 86.2 & 80.7 & 85.9 & 37.0 & 33.4 & 31.6 & 39.0 & 39.0 & 47.5 & 14.2 & 29.2 & 43.9 \\
 & Qwen3-4B & 3.23 & 4.28 & 3.25 & 2.88 & 2.69 & 3.27 & 78.8 & 74.4 & 79.7 & 26.0 & 28.5 & 22.4 & \underline{46.5} & \underline{46.5} & \underline{58.5} & 25.7 & 36.6 & 44.6 \\
 & Baichuan-M2-32B & \underline{3.81} & 4.74 & 4.08 & \underline{\textbf{3.73}} & \underline{3.87} & 4.05 & 80.0 & 73.3 & 80.0 & 38.5 & 37.6 & 38.0 & 42.5 & 42.5 & 55.5 & 25.2 & 34.9 & 47.0 \\
 & GPT-OSS-20B & \underline{\textbf{3.95}} & \underline{\textbf{4.95}} & \underline{\textbf{4.21}} & 3.57 & 3.79 & \underline{\textbf{4.09}} & 78.8 & 75.0 & 79.8 & 37.5 & 37.0 & 37.2 & 45.5 & 45.5 & 50.0 & 26.4 & 36.9 & 47.2 \\
 & Qwen3-32B & 3.74 & 4.67 & 3.97 & 3.47 & 3.48 & 3.87 & 73.8 & 71.4 & 75.5 & \underline{43.5} & \underline{\textbf{42.2}} & \underline{\textbf{45.7}} & 42.5 & 42.5 & 54.0 & 24.1 & 35.2 & 47.7 \\
\bottomrule
\end{tabular*}
\end{table*}

Dynamic results are summarized in Table~\ref{tab:dynamic_results}.
The 2-class accuracy can reach 92.3\% (DeepSeek-V3.2 under APA-Guided + MRD-RAG), but performance decreases substantially for 4-class and 12-class tasks.
The best 4-class accuracy is 43.0\% (Grok-4.1-Fast under APA-Guided + MRD-RAG), and the best 12-class accuracy is 28.5\% (Grok-4.1-Fast under APA-Guided + MRD-RAG).
Top-1 accuracy peaks at 37.5\% for 12-class prediction (Grok-4.1-Fast under APA-Guided + MRD-RAG), indicating that correct diagnoses are often present in candidates but remain difficult to rank as the primary label.
Overall, stronger consultation quality does not consistently translate into higher diagnostic accuracy, suggesting that diagnostic reasoning and interviewing skills need to be optimized separately.
We additionally run the dynamic evaluation with real patient profiles and dialogues from LingxiDiag-Clinical in place of LLM-simulated patients; results are reported in Table~\ref{tab:dynamic_real} and analyzed in Section~\ref{sec:conclusion}.

\section{Discussion and Conclusion}
\label{sec:conclusion}

We present LingxiDiagBench, a comprehensive benchmark for evaluating AI-assisted psychiatric diagnosis through agent-based consultation simulation.
LingxiDiagBench provides two complementary evaluation paradigms: static evaluation for reproducible dialogue analysis and dynamic evaluation for interactive consultation assessment.
The benchmark spans three difficulty levels from binary depression-anxiety classification to twelve-way differential diagnosis, enabling systematic assessment of diagnostic capabilities across varying clinical complexity.
Through extensive experiments encompassing state-of-the-art LLMs, we establish comprehensive performance baselines that reveal both the current capabilities and critical limitations of AI-assisted psychiatric diagnosis.

A fundamental challenge in building dynamic consultation benchmarks lies in the simulation environment itself, particularly the fidelity of Patient Agents.
Our evaluation demonstrates that the LingxiDiag-Patient Agents achieve Overall scores of 4.07--4.67 out of 5 (Table~\ref{tab:patient_eval}), with the best configuration (Qwen3-32B, 4.67) surpassing even the real clinical data baseline (4.30), whereas MDD-5K-Patient counterparts score only 2.23--4.09 under identical evaluation.
This performance gap underscores the importance of prompt design and domain-specific optimization for realistic patient simulation.
However, despite matching real data distributions, current Patient Agents still exhibit limitations in capturing the full diversity of individual patient presentations and communication styles.
Designing more personalized and authentic patient simulation environments that can serve as effective training grounds for Doctor Agents remains an open research challenge.

Our results reveal a substantial performance gap between static and dynamic evaluation paradigms, highlighting that static evaluation alone cannot fully capture the requirements of real-world consultation scenarios.
The diagnostic accuracy in dynamic settings often falls below that observed in static evaluation, indicating that ineffective information-gathering strategies can impair downstream diagnostic reasoning and reduce end-to-end performance.
This finding aligns with current reinforcement learning-based training paradigms that emphasize learning through interaction.
Notably, different consultation strategies (Free-form, Symptom-Tree, APA-Guided, and APA-Guided + MRD-RAG) yield varying diagnostic outcomes, demonstrating that how to ask questions is as important as what diagnostic conclusions to draw.
We observe that adding MRD-RAG to APA-Guided can improve end-to-end diagnostic performance on LingxiDiag-16K in the dynamic setting with an increase of 5\% in overall classification performance (Table~\ref{tab:dynamic_results}). 
We further conduct the dynamic evaluation using real patient profiles and dialogues from the LingxiDiag-Clinical dataset in place of LLM-simulated patients.
As shown in Table~\ref{tab:dynamic_real}, APA-Guided + MRD-RAG still yields the highest consultation quality (LLM-Ovl up to 4.09 for GPT-OSS-20B), consistent with the synthetic-data findings.
However, for diagnostic classification, the Symptom-Tree strategy with GPT-OSS-20B achieves the best Clf-Ovl of 50.0, outperforming APA-Guided + MRD-RAG, which differs from the synthetic-data setting, where APA-Guided + MRD-RAG leads.
Diagnostic accuracy on real patient data is generally higher (e.g., 2-class accuracy up to 91.2\%, 12-class accuracy up to 47.0\%), likely because real clinical cases present more prototypical symptom patterns.
This discrepancy between the two settings suggests that the benefit of retrieval-augmented strategies may be data-dependent, warranting further investigation.

\XSH{To validate the LLM-as-a-Judge evaluation, we provide a systematic comparison between LLM-as-Judge evaluations and licensed psychiatrist annotations, where 3-model AI ensemble (Gemma-3-27B, GPT-OSS-20B, and Qwen3-30B) and two independent licensed psychiatrists rate on 64 matched dialogue samples. Human ratings confirm that LingxiDiag-Patient significantly outperforms MDD-5K Patient across all six Patient dimensions (Mann--Whitney $U$ = 631.0, $p$ < 0.001, effect size $r$ = 0.547). For doctor agent evaluation, it demonstrates near-perfect concordance on strategy ranking: Kendall W = 0.90 and Spearman $\rho$ = 0.80 for Doctor Agent version ordering, with 83.3\% pairwise direction agreement under Mann-Whitney U tests (p-value<0.001). Full protocol and results are reported in Appendix~\ref{app:human_eval}.}

In the dynamic consultation evaluation, the LLM-as-a-Judge scores for dialogue quality exhibit an overall positive correlation with model scale, and proprietary models generally outperform open-source alternatives.
However, the correlation between LLM dialogue quality scores and classification task performance is moderate (r = 0.43), suggesting that high-quality consultation behavior does not automatically translate to accurate diagnostic outcomes.
This decoupling indicates substantial room for improvement in current models' diagnostic reasoning capabilities, even when their consultation conduct appears clinically appropriate.
Besides, integrated optimization of both consultation behavior and diagnostic decision-making, rather than relying solely on model scale, represents an important direction for advancing AI-assisted psychiatric diagnosis.
\XSH{Regarding computational efficiency, the multi-phase consultation design necessarily involves multiple LLM calls per session, reflecting the inherent complexity of structured clinical interviews rather than unnecessary overhead.
The framework provides several scaling mechanisms: the Doctor Agent under evaluation can be any model — including smaller or quantized variants — without framework modification; parallel API calls are natively supported to increase throughput; the Free-form strategy offers a lightweight single-prompt alternative suitable for rapid iterative testing; and the modular design allows individual components (e.g., static diagnosis only) to be evaluated independently without running the full dynamic pipeline.}

Several limitations should be considered when interpreting our findings.
First, although LingxiDiag-16K is constructed to match the distribution of a large real-world clinical dataset, it inevitably differs from authentic clinical encounters in certain aspects, with limited diversity in rare presentations and atypical symptom profiles.
\XSH{
Second, our benchmark is built on Chinese psychiatric consultation scenarios, with prompts in Chinese. Therefore, the reported results mainly reflect model performance in a Chinese linguistic and cultural context, and may not generalize to other languages or cultures due to data limitations. Nevertheless, the evaluation framework itself is language-agnostic. It is grounded in international clinical standards (e.g., APA/DSM-5), and both the EMR-based data generation pipeline and the multi-agent architecture are applicable across languages and LLMs. The evaluation dimensions are also universal clinical quality criteria. Thus, while current findings are Chinese-specific, the framework can be readily extended to cross-lingual settings.
}
Third, our current evaluation focuses on initial outpatient diagnosis and does not address treatment planning, follow-up assessment, or longitudinal care coordination; future work will extend the benchmark to cover these additional clinical workflow stages.

In conclusion, LingxiDiagBench establishes a standardized platform for benchmarking AI-assisted psychiatric diagnosis, combining real clinical data with large-scale synthetic cases to enable both authentic and scalable evaluation.
Our comprehensive experiments identify key challenges, including patient simulation fidelity, the gap between static and dynamic evaluation, comorbidity recognition, and the decoupling between consultation quality and diagnostic accuracy.
These resources aim to accelerate the development of AI systems that can ultimately improve access to accurate and timely psychiatric diagnosis while highlighting the current limitations that necessitate careful validation before clinical deployment.

\section{Ethical Use of Data}

\textbf{Data provenance and privacy.}
The clinical data are derived from de-identified electronic medical records collected at SMHC under institutional review board approval (IRB 2023-69) with signed consent forms for all participants.
All participant identifiers, including names, identification numbers, contact information, and precise timestamps, were removed prior to data processing.
The synthetic consultation dialogues in LingxiDiag-16K are generated entirely by LLMs based on statistical distributions extracted from the de-identified clinical data; no verbatim excerpts from real patient records are retained.
Two licensed psychiatrists reviewed random samples from the synthetic dataset and confirmed that no dialogue contains protected health information or content that could compromise patient privacy.

\textbf{Intended use and risk mitigation.}
LingxiDiagBench is designed exclusively for research purposes to advance AI-assisted psychiatric diagnosis and consultation modeling.
It is not a clinical diagnostic tool and must not be deployed in real clinical settings without rigorous validation, regulatory approval, and human oversight.
Synthetic dialogues may reflect biases inherited from the source clinical data and LLMs used in generation; users must conduct thorough bias audits and fairness evaluations before any downstream application.
We acknowledge potential risks, including inappropriate clinical adoption, algorithmic bias amplification, and the possibility that realistic synthetic dialogues could be misused to train systems without adequate safety guardrails.
We strongly emphasize that AI systems trained on this benchmark should augment, not replace, professional clinical judgment.


\bibliographystyle{ACM-Reference-Format}
\balance 
\bibliography{main}

\appendix

\XSH{
\section{Human Expert Validation of LLM-as-a-Judge}
\label{app:human_eval}

To validate the LLM-as-a-Judge evaluation pipeline, two licensed psychiatrists independently rated 64 matched dialogue samples spanning all 8 Doctor--Patient version combinations (4 Doctor strategies $\times$ 2 Patient versions).
Each dialogue was scored on the 5 Doctor Agent dimensions (Clinical Accuracy \& Competence, Ethical and Professional Conduct, Assessment and Response, Therapeutic Relationship and Alliance, AI Communication Quality) and 6 Patient Agent dimensions (Accuracy, Honesty, Brevity, Proactivity, Restraint, and Polish) under the same rating rules and scales as the LLM-as-a-Judge rating scales.

First, human ratings confirm that LingxiDiag-Patient significantly outperforms MDD-5K Patient across all six Patient dimensions (Mann--Whitney $U$ = 631.0, $p$ < 0.001, effect size $r$ = 0.547). The largest differences appear in Response Brevity (+1.448), Emotional Restraint (+1.385), and Information Proactivity (+1.219), all reaching $p$ < 0.001. 
Moreover, using rank-based concordance metrics appropriate for raters with different score calibrations, we found near-perfect agreement on the Doctor strategy ranking (Kendall W = 0.90, Spearman $\rho$ = 0.80), with both methods identifying APA-Guided as the best strategy and Free-form as the worst. Mann-Whitney U pairwise comparisons confirmed 83.3\% direction agreement.

\section{Cross-Dataset Validation}
\label{app:sft_transfer}

To provide direct evidence that LingxiDiag-16K captures clinically realistic patterns beyond surface-level statistics, we fine-tuned Qwen3-8B and Qwen3-32B using LoRA-based supervised SFT on the LingxiDiag-16K training split, where the LoRA rank, learning rate, and epoch are set as 32, 5e-5, and 3, respectively, and evaluated on both the LingxiDiag-16K validation set (synthetic) and the LingxiDiag-Clinical validation set (real clinical data).
If the synthetic data merely preserved surface-level distributional properties, knowledge learned from it would not be expected to generalize to real clinical scenarios.

Table~\ref{tab:sft_transfer} summarizes the results.
Fine-tuning on LingxiDiag-16K yields consistent and substantial improvements on real clinical data, most prominently in the 12-class diagnostic task:
Qwen3-8B improves from 4.1\% to 41.4\% exact-match 12-class accuracy (+37.3\%) on LingxiDiag-Clinical, and Qwen3-32B improves from 20.4\% to 39.7\% (+19.3\%).
The overall classification score (Clf-Ovl) likewise improves from 0.529 to 0.553 for Qwen3-8B and from 0.548 to 0.558 for Qwen3-32B on real data.
This cross-dataset transfer demonstrates that LingxiDiag-16K encodes semantically meaningful clinical knowledge that generalizes to real-world psychiatric consultations, supporting its utility as a training resource even when the deployment target is real clinical data.

\begin{table*}[htbp]
\centering
\footnotesize
\caption{Cross-dataset transfer results. Models are fine-tuned on the LingxiDiag-16K (synthetic) training split and evaluated on both the synthetic test set (top) and the real LingxiDiag-Clinical test set (bottom). Best results per section are \underline{\textbf{bold with underline}}; second-best are \underline{underlined}.}
\label{tab:sft_transfer}
\setlength\tabcolsep{2pt}
\begin{tabular*}{\linewidth}{@{\extracolsep{\fill}}llccccccccccccc@{}}
\toprule
& & \multicolumn{3}{c}{2-class} & \multicolumn{3}{c}{4-class} & \multicolumn{5}{c}{12-class} & \\
\cmidrule(lr){3-5} \cmidrule(lr){6-8} \cmidrule(lr){9-13}
Test Set & Model & Acc & m-F1 & w-F1 & Acc & m-F1 & w-F1 & Acc & Top1 & Top3 & m-F1 & w-F1 & Overall \\
\midrule
\multirow{4}{*}{\makecell{LingxiDiag\\-16K}}
 & Qwen3-8B            & 0.835 & 0.776 & 0.824 & 0.408 & 0.335 & 0.355 & 0.012 & 0.459 & \underline{\textbf{0.599}} & \underline{0.177} & \underline{0.420} & 0.473 \\
 & Qwen3-8B + LoRA-SFT & \underline{\textbf{0.846}} & \underline{\textbf{0.797}} & \underline{\textbf{0.839}} & 0.421 & 0.326 & 0.336 & \underline{0.391} & \underline{0.486} & 0.497 & 0.161 & 0.396 & 0.500 \\
\cmidrule(lr){2-14}
 & Qwen3-32B            & 0.827 & 0.791 & 0.827 & \underline{\textbf{0.438}} & \underline{\textbf{0.384}} & \underline{\textbf{0.404}} & 0.241 & 0.470 & \underline{0.566} & \underline{\textbf{0.188}} & \underline{\textbf{0.431}} & \underline{0.506} \\
 & Qwen3-32B + LoRA-SFT & \underline{0.839} & \underline{0.793} & \underline{0.834} & \underline{0.436} & \underline{0.362} & \underline{0.379} & \underline{\textbf{0.422}} & \underline{\textbf{0.496}} & 0.513 & \underline{0.177} & 0.417 & \underline{\textbf{0.515}} \\
\midrule
\multirow{4}{*}{\makecell{LingxiDiag\\-Clinical}}
 & Qwen3-8B            & 0.851 & \underline{0.808} & \underline{0.854} & 0.456 & 0.451 & 0.446 & 0.041 & \underline{\textbf{0.512}} & \underline{\textbf{0.669}} & 0.256 & \underline{\textbf{0.474}} & 0.529 \\
 & Qwen3-8B + LoRA-SFT & \underline{\textbf{0.862}} & \underline{\textbf{0.813}} & \underline{\textbf{0.861}} & 0.494 & 0.429 & 0.441 & \underline{\textbf{0.414}} & \underline{0.511} & 0.533 & \underline{0.277} & 0.444 & \underline{0.553} \\
\cmidrule(lr){2-14}
 & Qwen3-32B            & 0.824 & 0.780 & 0.829 & \underline{\textbf{0.524}} & \underline{\textbf{0.526}} & \underline{\textbf{0.523}} & 0.204 & 0.472 & \underline{0.601} & \underline{\textbf{0.278}} & \underline{0.470} & 0.548 \\
 & Qwen3-32B + LoRA-SFT & \underline{0.852} & 0.801 & 0.852 & \underline{0.506} & \underline{0.475} & \underline{0.485} & \underline{0.397} & 0.504 & 0.543 & 0.265 & 0.458 & \underline{\textbf{0.558}} \\
\bottomrule
\end{tabular*}
\end{table*}

\section{Synthesis Pipeline Details}
\label{app:synthesis}

The EMR synthesis pipeline generates synthetic Electronic Medical Records (EMRs) that preserve the demographic and clinical distributions of the LingxiDiag-Clinical dataset.
The pipeline proceeds in seven steps:

\begin{enumerate}
\setlength\itemsep{0em}
\item \textbf{Basic Information Sampling:} Age, gender, department, and ICD-10 diagnosis code are independently sampled from their respective empirical distributions extracted from the real clinical data.
\item \textbf{Accompanying Person Generation:} Presence and relationship of an accompanying person are sampled from age- and gender-conditioned distributions (e.g., younger patients are more likely to be accompanied by a parent).
\item \textbf{Personal History Generation:} Personal history fields---pregnancy status, developmental status, marital status, occupation, menstrual status (females only), personality traits, and special habits---are sampled from age-grouped distributions (0--18, 18--30, 30--45, 45--60, 60+) to capture realistic life-course patterns.
\item \textbf{Chief Complaint Generation:} Symptoms are sampled from diagnosis-specific symptom distributions (2--3 symptoms per case), combined with sampled duration expressions and target lengths drawn from diagnosis-specific text length distributions. An optional LLM polishing step refines the generated text for naturalness.
\item \textbf{Present Illness History Generation:} Trigger events and clinical keywords are sampled from diagnosis-conditioned distributions, then composed into coherent present illness narratives via LLM generation with target length constraints.
\item \textbf{Auxiliary Field Generation:} Physical illness history, drug allergy history, and family psychiatric history are sampled from their respective population-level distributions.
\item \textbf{EMR Assembly:} All generated fields are combined into a complete EMR record with a unique patient identifier, ICD-10 diagnosis code, and diagnosis results.
\end{enumerate}

All distribution mappings are derived from statistical analysis of the 1,709 real clinical cases and stored as structured JSON files.
The pipeline supports both sequential and parallel batch generation via thread pooling.

\section{Retrieval Procedure Details}
\label{app:retrieval}

The MRD-RAG variant of APA-Guided consultation augments the Doctor Agent with a retrieval-augmented generation module that provides relevant diagnostic guidelines during the Assessment and Deep-Dive consultation phases.

\textbf{Knowledge Base Construction.}
A Chinese clinical guideline document is loaded and split into text chunks of 500 characters with 50-character overlap, using sentence-boundary-aware splitting (splitting at Chinese sentence terminators).
Each chunk is embedded using a multilingual embedding model (Qwen3-Embedding-8B) and indexed in a FAISS \texttt{IndexFlatIP} index for inner-product similarity search (equivalent to cosine similarity with L2-normalized embeddings).

\textbf{Retrieval Process.}
During consultation, the Doctor Agent formulates a query based on the current dialogue context and suspected diagnoses.
The query is encoded using the same embedding model, and the top-$k$ ($k$ = 5) most similar chunks are retrieved from the FAISS index.
An optional re-ranking step using a cross-encoder model (Qwen3-Reranker-8B) refines the results to the top 3 most relevant passages.

\textbf{Integration with Consultation.}
Retrieved guideline passages are injected into the Doctor Agent's prompt context during the Assessment and Deep-Dive phases only, providing evidence-based diagnostic criteria and recommended follow-up questions for the top 3 suspected diagnoses.
The Doctor Agent then generates its next question, informed by both the dialogue history and the retrieved clinical knowledge.

\section{Metric Definitions}
\label{app:metrics}

\subsection{Classification Metrics}

For single-label classification tasks (2-class and 4-class), we report:
\begin{itemize}
\setlength\itemsep{0em}
\item \textbf{Accuracy:} Fraction of correctly classified samples.
\item \textbf{Macro F1 (m-F1):} Unweighted average of per-class F1 scores, treating all classes equally regardless of support.
\item \textbf{Weighted F1 (w-F1):} Average of per-class F1 scores weighted by class support, accounting for class imbalance.
\end{itemize}

For the 12-class multi-label classification task, we additionally report:
\begin{itemize}
\setlength\itemsep{0em}
\item \textbf{Top-1 Accuracy:} Fraction of samples where the first predicted label matches any ground-truth label.
\item \textbf{Top-3 Accuracy:} Fraction of samples where at least one of the top 3 predicted labels matches any ground-truth label.
\end{itemize}

Per-class precision, recall, and F1 are computed using scikit-learn with \texttt{zero\_division=0}.

\subsection{Generation Metrics}

For the next-question prediction task, we report:
\begin{itemize}
\setlength\itemsep{0em}
\item \textbf{BLEU:} Multi-gram precision with brevity penalty, using character-level tokenization for Chinese text.
\item \textbf{ROUGE-L:} F-measure based on the longest common subsequence between prediction and reference.
\item \textbf{BERTScore:} Semantic similarity computed using pre-trained BERT embeddings.
\end{itemize}

\begin{figure*}[htbp]
\centering
\includegraphics[width=0.65\textwidth]{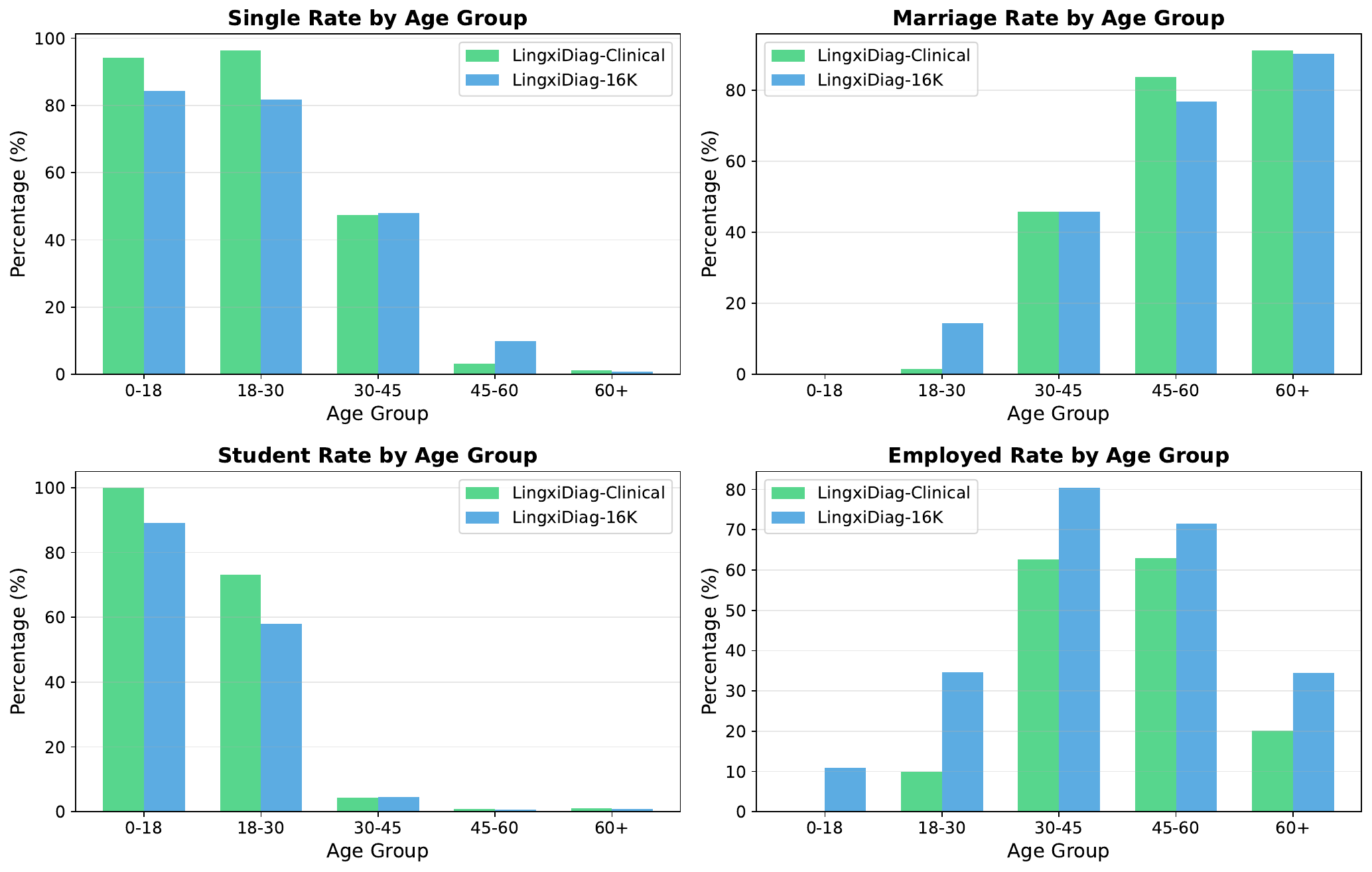}
\caption{Comparison of personal history distributions between LingxiDiag-Clinical and LingxiDiag-16K across age groups.}
\label{fig:personal_history}
\end{figure*}

\begin{figure*}[htbp]
\centering
\includegraphics[width=0.8\textwidth]{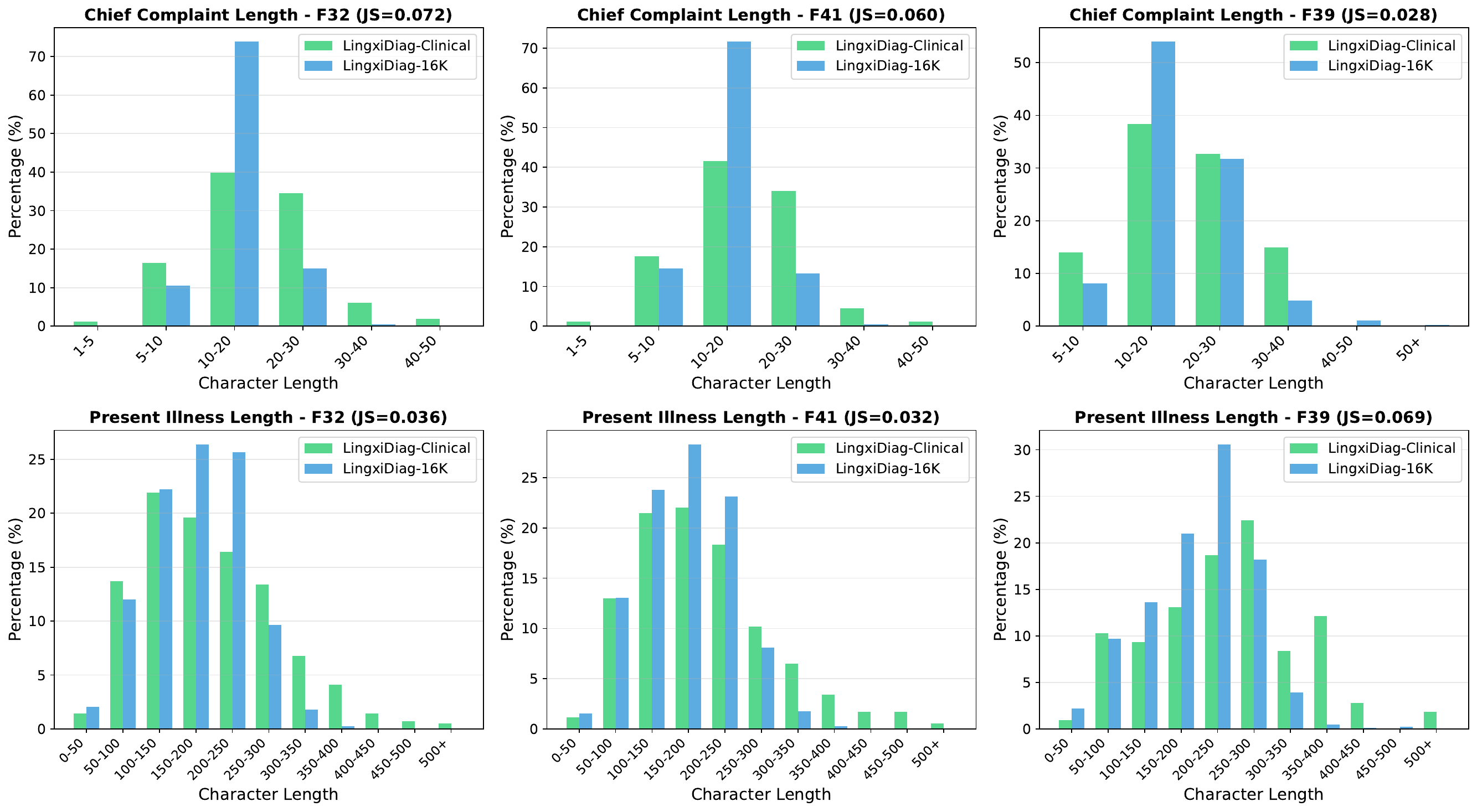}
\caption{Text length distribution comparison for chief complaints and present illness history across diagnostic categories (F32, F41, F39). Jensen-Shannon (JS) divergence values indicate high distributional similarity between LingxiDiag-Clinical and LingxiDiag-16K dataset.}
\label{fig:chief_complaint}
\end{figure*}

\subsection{LLM-as-a-Judge Scoring}

Both Patient Agent and Doctor Agent evaluations use a multi-model ensemble protocol.
Three evaluator models (Gemma-3-27B, GPT-OSS-20B, Qwen3-30B) independently score each dialogue on the respective evaluation dimensions.
For any missing score, the median of the remaining models for that sample and dimension is used as imputation.
The final score is the arithmetic mean across the three models for each dimension:
$$\text{Score}_{d} = \frac{1}{3}\sum_{m=1}^{3} s'_{m,d}$$
where $s'_{m,d}$ is the (possibly imputed) score from model $m$ on dimension $d$.
The Overall score is the arithmetic mean across all dimensions.
Doctor Agent evaluation uses a 1--6 Likert scale across 5 dimensions; Patient Agent evaluation uses a 1--5 scale across 6 dimensions.

}

\end{document}